\documentclass{article}

\PassOptionsToPackage{numbers}{natbib}

\usepackage[preprint]{neurips_data_2021}

\usepackage[utf8]{inputenc} 
\usepackage[T1]{fontenc}    
\usepackage{hyperref}       
\usepackage{url}            
\usepackage{booktabs}       
\usepackage{amsmath}
\usepackage{amsfonts}       
\usepackage{nicefrac}       
\usepackage{microtype}      
\usepackage{xcolor}         
\usepackage{graphics}
\usepackage{graphicx}
\usepackage{wrapfig}
\usepackage{listings}

\definecolor{dkred}{rgb}{0.8,0,0}
\definecolor{dkgreen}{rgb}{0,0.4,0}
\definecolor{gray}{rgb}{0.2,0.2,0.2}
\definecolor{mauve}{rgb}{0.7,0,0.9}
\title{SCOPE-RL: A Python Library for Offline Reinforcement Learning and Off-Policy Evaluation}

\author{%
  Haruka Kiyohara\thanks{This work was done during their internship at negocia, Inc.} \\
  Cornell University \\
  \texttt{hk844@cornell.edu} \\
  \And
  Ren Kishimoto$^{\ast}$ \\
  Tokyo Institute of Technology \\
  \texttt{kishimoto.r.ab@m.titech.ac.jp} \\
  \AND
  Kosuke Kawakami\\
  HAKUHODO Technologies Inc.\\
  \texttt{kosuke\_kawakami@negocia.jp} \\
  \And
  Ken Kobayashi\\
  Tokyo Institute of Technology\\
  \texttt{kobayashi.k.ar@m.titech.ac.jp} \\
  \And
  Kazuhide Nakata\\
  Tokyo Institute of Technology\\
  \texttt{nakata.k.ac@m.titech.ac.jp} \\
  \And
  Yuta Saito\\
  Cornell University\\
  \texttt{ys552@cornell.edu} \\
}

\usepackage{algorithm}
\usepackage{algorithmic}
\usepackage{bbding}

\usepackage{multicol,multirow}
\usepackage{listings}
\definecolor{dkgreen}{rgb}{0,0.6,0}
\definecolor{customgray}{rgb}{0.25,0.25,0.25}
\definecolor{customred}{rgb}{0.8,0.05,0.05}
\definecolor{customblue}{rgb}{0.05,0.05,0.8}
\usepackage{bbding}
\usepackage{comment}
\usepackage{bm}
\usepackage{tcolorbox}

\newcommand{\mE}{\mathbb{E}}

\newcommand{\calD}{\mathcal{D}}

\newcommand{\calA}{\mathcal{A}}

\definecolor{dkred}{rgb}{0.8,0,0}
\definecolor{dkpink}{rgb}{1.0,0,0.5}
\definecolor{dkgreen}{rgb}{0,0.4,0}
\definecolor{tickgreen}{rgb}{0,0.6,0}
\newcommand{\tick}{\textcolor{tickgreen}{\CheckmarkBold}}
\newcommand{\fail}{\textcolor{red}{\XSolidBrush}}

\begin{document}

\maketitle

\begin{figure}[H]
  \centering
  \includegraphics[clip, width=0.6\linewidth]{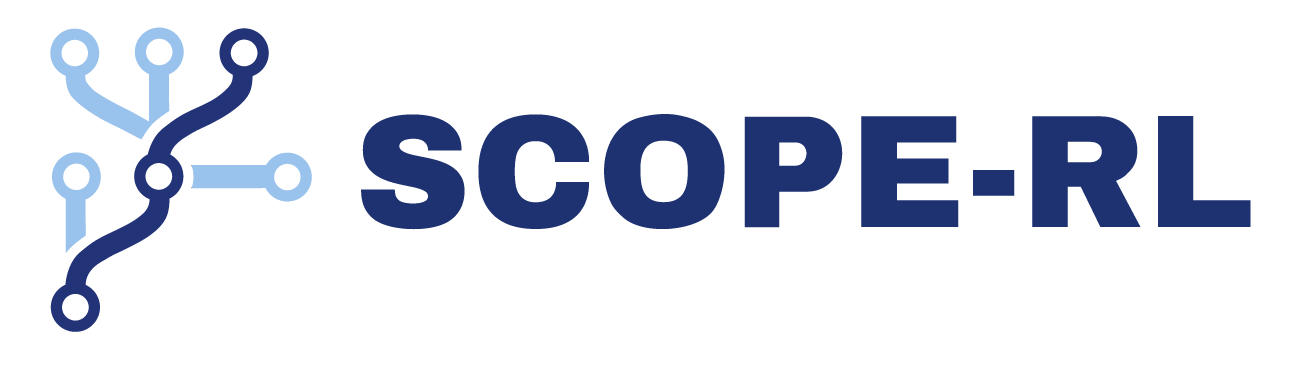}
\end{figure}

\begin{abstract}
This paper introduces \textbf{SCOPE-RL}, a comprehensive open-source Python software designed for offline reinforcement learning (offline RL), off-policy evaluation (OPE), and selection (OPS). Unlike most existing libraries that focus solely on either policy learning or evaluation, SCOPE-RL seamlessly integrates these two key aspects, facilitating flexible and complete implementations of both offline RL and OPE processes. SCOPE-RL put particular emphasis on its OPE modules, offering a range of OPE estimators and robust evaluation-of-OPE protocols. This approach enables more in-depth and reliable OPE compared to other packages. For instance, SCOPE-RL enhances OPE by estimating the entire reward distribution under a policy rather than its mere point-wise expected value. Additionally, SCOPE-RL provides a more thorough evaluation-of-OPE by presenting the risk-return tradeoff in OPE results, extending beyond mere accuracy evaluations in existing OPE literature. SCOPE-RL is designed with user accessibility in mind. Its user-friendly APIs, comprehensive documentation, and a variety of easy-to-follow examples assist researchers and practitioners in efficiently implementing and experimenting with various offline RL methods and OPE estimators, tailored to their specific problem contexts. The documentation of SCOPE-RL is available at \textcolor{dkpink}{\textbf{\href{https://scope-rl.readthedocs.io/en/latest/}{https://scope-rl.readthedocs.io/en/latest/}}}.
\end{abstract}

\section{Introduction}

\begin{table}
  \caption{Comparing SCOPE-RL with existing offline RL and OPE packages}
  \label{tab:scope_rl}
  \centering
  \scalebox{0.850}{
  \begin{tabular}{c||c|c|cc|c}
    \toprule
     & data collection & offline RL & OPE & CD-OPE & evaluation of OPE \\
    \midrule
    offline RL packages$^{\ast 1}$  & \tick & \tick & (limited) & \fail & \fail \\
    application-specific test beds$^{\ast 2}$ & (limited) & \tick & (limited) & \fail & \fail \\
    \midrule
    DOPE~\citep{fu2021benchmarks} & (limited) & (limited) & \tick & \fail & (w/o SharpRatio@k) \\
    COBS~\citep{voloshin2019empirical} & (limited) & \fail & \tick & \fail & (w/o SharpRatio@k) \\
    OBP~\citep{saito2021open} & (non-RL) & \fail & (non-RL) & \fail & (non-RL) \\
    \midrule 
    \textbf{SCOPE-RL (ours)} & \tick & \tick & \tick & \tick & \tick \\
    \bottomrule
  \end{tabular}
  }
  \vskip 0.1in
  \raggedright
  \fontsize{9pt}{9pt}\selectfont \textit{Note}:
  In the column ``data collection'', \tick means that the package is compatible with Gym/Gymnasium~\citep{brockman2016openai} environments and thus is able to handle various simulation settings. In the column ``offline RL'', \tick means that the package implements a variety of offline RL algorithms or is compatible to one of offline RL libraries. In particular, our SCOPE-RL supports compatibility with d3rlpy~\citep{seno2021d3rlpy}. In the column ``OPE'', \tick means that the package implements various OPE estimators other than standard choices such as Direct Method~\citep{le2019batch}, Importance Sampling~\citep{precup2000eligibility}, and Doubly Robust~\citep{jiang2016doubly}. (limited) means that the package supports only these standard estimators. CD-OPE is the abbreviation of Cumulative Distribution OPE, which estimates the cumulative distribution function of the return under evaluation policy~\citep{chandak2021universal,huang2022off}. Note that ``offline RL packages$^{\ast 1}$'' refers to d3rlpy~\citep{seno2021d3rlpy}, CORL~\citep{tarasov2024corl}, RLlib~\citep{liang2018rllib}, and Horizon~\citep{gauci2018horizon}, while ``application-specific testbeds$^{\ast 2}$'' refers to NeoRL~\citep{qin2021neorl}, RecoGym~\citep{rohde2018recogym}, RL4RS~\citep{wang2021rl4rs}, and AuctionGym~\citep{jeunen2023off}.
\end{table}

Reinforcement learning (RL) has garnered significant interest in numerous sequential decision-making scenarios, such as healthcare, education, recommender systems, and robotics. However, its online learning process is often deemed impractical for real-world applications due to the costly and potentially harmful nature of active exploration in the environment~\citep{fu2021benchmarks,kiyohara2021accelerating,matsushima2021deployment,saito2021evaluating}. To overcome these challenges, learning and evaluating new policies offline from existing historical data, known as offline RL~\citep{levine2020offline} and off-policy evaluation (OPE)~\citep{fu2021benchmarks}, have become increasingly prevalent approaches for applying RL in real-world scenarios~\citep{kiyohara2023towards,kurenkov2022showing, qin2021neorl}.

While policy learning and evaluation are both vital in the offline RL process, current packages typically focus on only one of these aspects, lacking the flexibility to integrate both seamlessly. Most offline RL libraries~\citep{gauci2018horizon,liang2018rllib,qin2021neorl,seno2021d3rlpy,tarasov2022corl} emphasize policy learning, offering limited OPE estimators for policy evaluation. Further, these packages generally lack comprehensive evaluation protocols for OPE, which are essential for benchmarking and advancing new OPE estimators. In contrast, existing packages dedicated to OPE, such as DOPE~\citep{fu2021benchmarks} and COBS~\citep{voloshin2019empirical}, while providing valuable testbeds for OPE, are not as adaptable in accommodating a variety of environments and offline RL methods.

Driven by the limited availability of offline RL packages that effectively integrate both policy learning and evaluation, we introduce \textbf{SCOPE-RL}, the first comprehensive software designed to streamline the entire offline RL-to-OPE process, available at \textcolor{dkpink}{\textbf{\href{https://github.com/hakuhodo-technologies/scope-rl}{https://github.com/hakuhodo-technologies/scope-rl}}}. SCOPE-RL, developed in Python, stands out for its two-fold focus. Firstly, unlike most offline RL packages, SCOPE-RL places a strong emphasis on its OPE modules, incorporating a range of OPE estimators (as described in Sections~\ref{sec:implemented_ope} and \ref{sec:cumulative_distribution_ope}) and their comprehensive assessment protocols (as described in Section~\ref{sec:assessments}). These features enable practitioners to conduct thorough policy evaluations and researchers to carry out more insightful evaluation of OPE estimators than those possible with existing ones. For instance, SCOPE-RL allows for the estimation of a policy's performance distribution in addition to the usual point-wise estimate, i.e., \textit{cumulative distribution OPE}~\citep{chandak2021universal,huang2021off,huang2022off}. Our software also facilitates evaluation-of-OPE based on the risk-return tradeoff, not just the accuracy of OPE and downstream policy selection tasks~\citep{kiyohara2023towards}. Secondly, SCOPE-RL extends beyond mere OPE libraries by supporting compatibility with OpenAI Gym/Gymnasium environments~\citep{brockman2016openai} and d3rlpy~\citep{seno2021d3rlpy}, which implements various offline RL algorithms. This extension ensures that SCOPE-RL can provide flexible, end-to-end solutions for offline RL and OPE across various environments and methods. In addition, our user-friendly APIs, visualization tools, detailed documentation, and diverse quickstart examples ease the implementation of offline RL and OPE in a range of problem settings, as demonstrated in Appendix~\ref{app:example}.

To summarize, the key features of SCOPE-RL include:
\begin{itemize}
    \item \textbf{End-to-end implementation of offline RL and OPE:} SCOPE-RL facilitates a seamless process from data collection through offline RL, OPE, and up to the assessment of OPE, particularly focusing on OPE modules and compatibility with Gym/Gymnasium~\citep{brockman2016openai} and d3rlpy~\citep{seno2021d3rlpy}. (Section \ref{sec:overview})
    \item \textbf{Variety of OPE estimators:} SCOPE-RL incorporates not only basic OPE estimators~\citep{jiang2016doubly, le2019batch, precup2000eligibility, thomas2015confidence} but also advanced ones such as marginal OPE estimators~\citep{kallus2020double, liu2018breaking, uehara2020minimax,  yuan2021sope}, high confidence OPE~\citep{thomas2015confidence, thomas2015high}, and cumulative distribution OPE~\citep{chandak2021universal,huang2021off,huang2022off}. (Section \ref{sec:implemented_ope})
    \item \textbf{Cumulative distribution OPE for risk function estimation:} SCOPE-RL is the first to implement cumulative distribution OPE~\citep{chandak2021universal,huang2021off,huang2022off}, estimating the performance distribution of a policy and various risk functions like variance and conditional value-at-risk. (Section \ref{sec:cumulative_distribution_ope})
    \item \textbf{Risk-return assessments of OPE and downstream policy selection:} SCOPE-RL excels in implementing many evaluation-of-OPE metrics including the one to assess the risk-return tradeoff in downstream policy selection tasks~\citep{kiyohara2023towards}. (Section \ref{sec:assessments})
    \item \textbf{User-friendly APIs, visualization tools, and documentation:} SCOPE-RL enhances ease of use with its intuitive API design, comprehensive documentation, and an array of quickstart examples. (Section \ref{sec:api})
\end{itemize}

Table~\ref{tab:scope_rl} offers an in-depth comparison of SCOPE-RL with other existing packages.

\section{Overview of SCOPE-RL} \label{sec:overview}

\begin{figure}
  \centering
  \includegraphics[clip, width=0.90\linewidth]{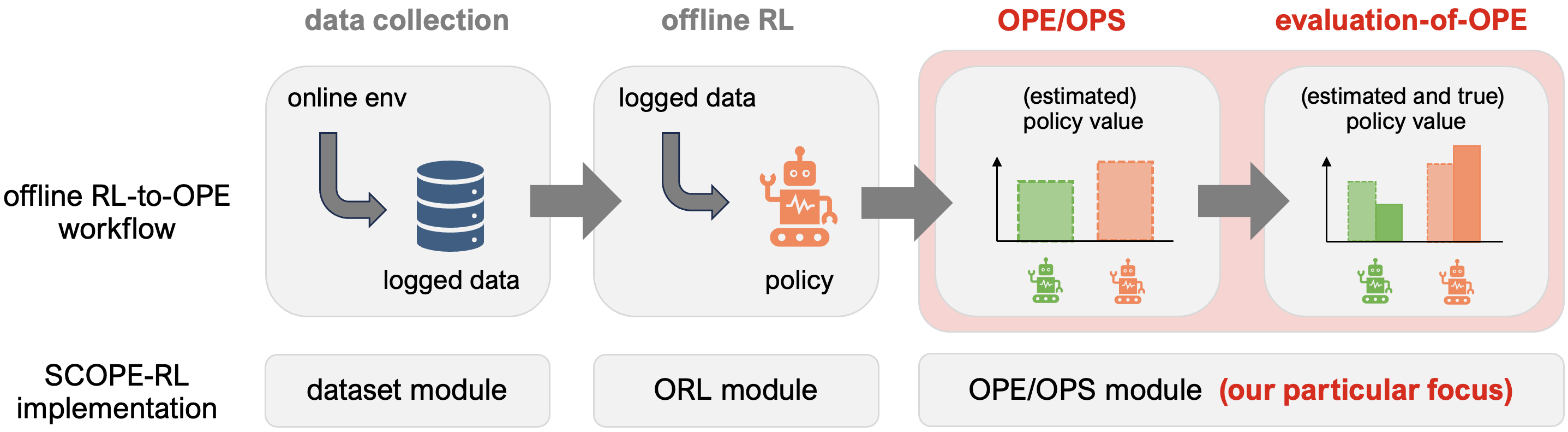}
  \caption{End-to-end workflow of offline RL and OPE with SCOPE-RL.} \label{fig:workflow}
\end{figure}

While existing packages offer flexible implementations for either offline RL or OPE, none currently provide a comprehensive, end-to-end solution that encompasses offline RL, OPE, and evaluation-of-OPE metrics. SCOPE-RL bridges this gap, seamlessly integrating the implementation of offline RL and OPE processes in an end-to-end manner for the first time. Specifically, to streamline the implementation process, our package comprises four key modules, with a particular focus on the latter two, as depicted in Figure~\ref{fig:workflow}:
\begin{tcolorbox}[boxrule = 0.5pt, colframe = white,]
\underline{\textbf{SCOPE-RL Modules}}
\normalsize
\begin{itemize}
    \item Dataset module
    \item Offline Policy Learning (Offline Reinforcement Learning; ORL) module
    \item \textbf{Off-Policy Evaluation (OPE) module}
    \item \textbf{Off-Policy Selection (OPS) module}
\end{itemize}
\end{tcolorbox}
The ``Dataset'' module is responsible for data collection and generation from RL environments. Thanks to its compatibility with OpenAI Gym/Gymnasium~\citep{brockman2016openai}-like environments, SCOPE-RL can be applied to a wide range of environmental settings. Furthermore, SCOPE-RL's compatibility with d3rlpy~\citep{seno2021d3rlpy}, which includes various online and offline RL algorithms, allows users to assess the effectiveness of offline RL algorithms and OPE estimators across diverse data collection policies and experimental configurations.

The ``ORL'' module in SCOPE-RL offers a user-friendly wrapper for developing new policies using various offline RL algorithms. While d3rlpy~\citep{seno2021d3rlpy} already features an accessible API, it is primarily geared towards employing offline RL algorithms individually. To enhance the efficiency of the entire offline RL and OPE process, our ORL module facilitates the management of multiple datasets and algorithms within a single unified class as explained in Appendix~\ref{app:example_multiple} in greater detail.

The core of SCOPE-RL lies in the ``OPE'' and ``OPS'' modules. As elaborated in the following sections, we have incorporated a diverse array of OPE estimators in SCOPE-RL, ranging from basic options~\citep{jiang2016doubly, le2019batch, precup2000eligibility, thomas2016data} to advanced estimators that use marginal importance sampling~\citep{kallus2020double, liu2018breaking, uehara2020minimax, yang2020off, yuan2021sope}, and those tailored for cumulative distribution OPE~\citep{chandak2021universal,huang2021off,huang2022off}. Additionally, we include various evaluation-of-OPE metrics. These key features in SCOPE-RL allow for a more nuanced understanding of policy performance and the effectiveness of OPE estimators, such as estimating a policy's performance distribution (CD-OPE), and evaluating OPE outcomes in terms of the risk-return tradeoff in downstream policy selection tasks, as illustrated in Figure~\ref{fig:ope_features}.

SCOPE-RL also introduces meta-classes for managing OPE/OPS experiments and abstract base classes for implementing new OPE estimators. These features enable researchers to rapidly integrate and test their own algorithms within the SCOPE-RL framework, and aid practitioners in comprehending the characteristics of various OPE methods through empirical evaluation.

\section{Implemented OPE estimators and evaluation-of-OPE metrics} \label{sec:implemented_ope}

\begin{figure}
  \centering
  \includegraphics[clip, width=0.90\linewidth]{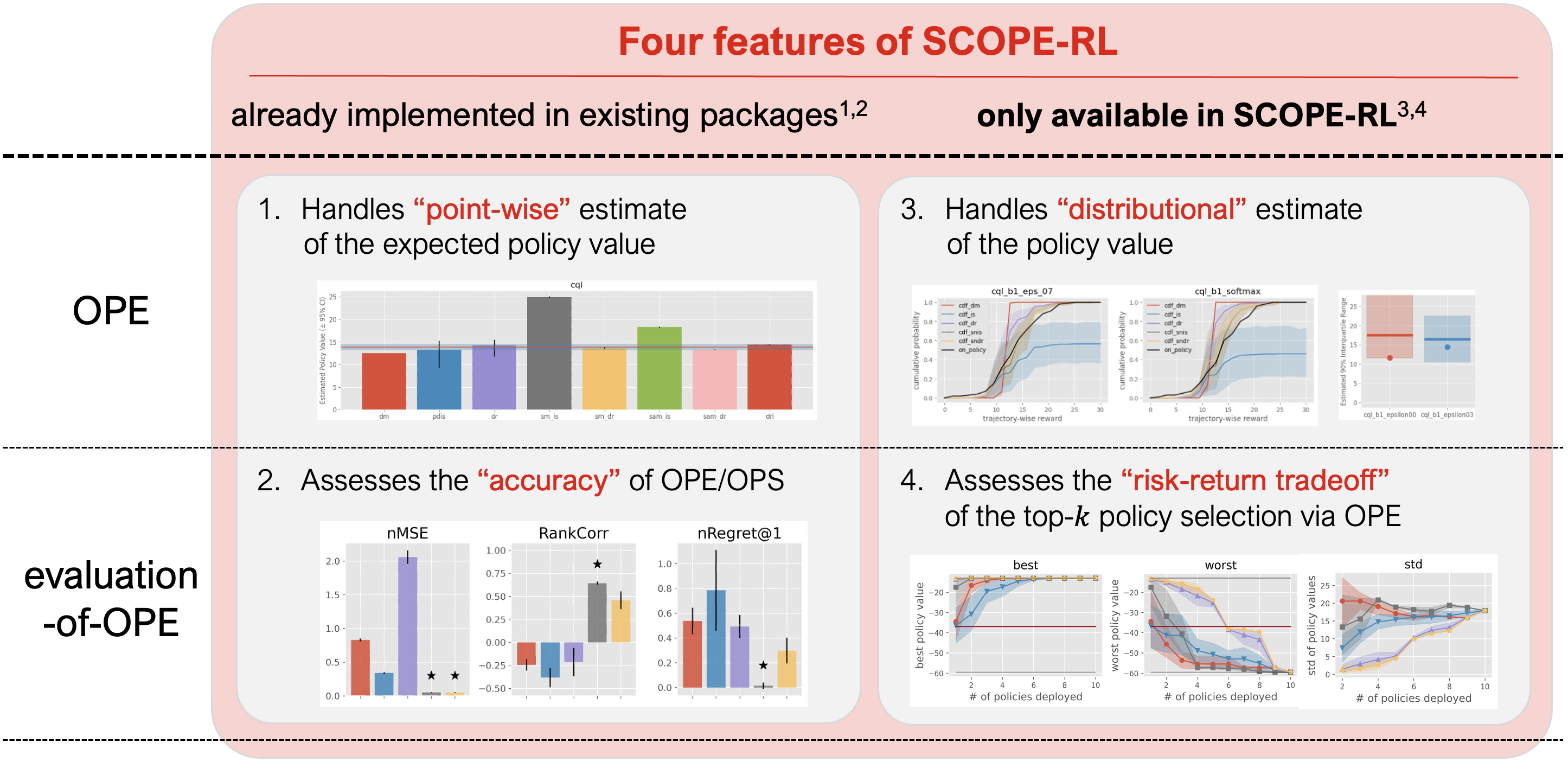}
  \caption{
  Summarizing the distinctive features of SCOPE-RL.
  \textbf{OPE}: While existing packages (e.g., \citep{fu2021benchmarks, voloshin2019empirical}) focus only on estimating the expected performance in a point-wise manner (left), SCOPE-RL additionally supports \textit{cumulative distribution OPE}~\citep{chandak2021universal, huang2021off, huang2022off} to estimate the whole distribution of policy performance (right). \textbf{Evaluation-of-OPE}: While existing package reports only the ``accuracy'' of OPE or that of the downstream policy selection tasks~\citep{kiyohara2023towards} (left), SCOPE-RL also measures various risk-return tradeoff metrics in top-$k$ policy selection (right) (See Section \ref{sec:assessments} for the details). \textbf{Visualization}:  Finally, all figures, including those illustrating the properties of existing packages, are generated by the visualization tools implemented in SCOPE-RL.
  } \label{fig:ope_features}
\end{figure}

A distinctive contribution of SCOPE-RL is its comprehensive suite of OPE estimators. SCOPE-RL not only includes foundational OPE estimators like Fitted Q-Evaluation~\citep{le2019batch}, Per-Decision Importance Sampling~\citep{precup2000eligibility}, and Doubly Robust~\citep{jiang2016doubly, thomas2016data}, but it also integrates advanced estimators that use state or state-action marginal importance weights~\citep{kallus2020double, liu2018breaking, uehara2020minimax, yuan2021sope}, high-confidence OPE~\citep{thomas2015confidence, thomas2015high}, and cumulative distribution OPE~\citep{chandak2021universal, huang2021off, huang2022off}, alongside unique evaluation metrics for OPE~\citep{kiyohara2023towards}.

In particular, the cumulative distribution OPE and the novel evaluation-of-OPE metrics set SCOPE-RL apart from existing OPE packages like \citep{fu2021benchmarks, le2019batch}. Through cumulative distribution OPE, SCOPE-RL is capable of estimating the entire performance distribution of a policy, in contrast to traditional OPE methods that only compute the point-wise expected policy performance~\citep{chandak2021universal, huang2021off, huang2022off}. Furthermore, our evaluation-of-OPE metrics, based on the statistics of the top-$k$ policies selected by OPE (``policy portfolio''), offer insights into the risk-return tradeoff in policy selection~\citep{kiyohara2023towards}. This approach transcends the conventional metrics such as Mean-Squared Error (MSE), Rank Correlation (RankCorr), and Regret, which focus only on ``accuracy'' in OPE and downstream policy selection. Consequently, SCOPE-RL enables a more multifaceted comparison of policy performance and the efficacy of OPE estimators compared to existing packages, as illustrated in Figure~\ref{fig:ope_features}. We will delve into these key features of SCOPE-RL in greater detail in Sections~\ref{sec:cumulative_distribution_ope} and \ref{sec:assessments}.

The following is an overview of the OPE estimators and evaluation-of-OPE metrics implemented in SCOPE-RL. For a more comprehensive and rigorous understanding of each estimator’s definition and properties, please refer to Appendix~\ref{app:implementation}.

\vspace{2mm}

\begin{tcolorbox}[colback=green!12!white,  boxrule = 0.5pt, colframe = white,]
\large{\underline{\textbf{Standard Off-Policy Evaluation (OPE)}}}
\normalsize
\begin{itemize}
    \item \textbf{Basic Estimators}:
    \begin{itemize}
        \item (abstract base implementation)
        \item Direct Method (DM)~\citep{le2019batch}
        \item Trajectory-wise Importance Sampling (TIS)~\citep{precup2000eligibility}
        \item Per-Decision Importance Sampling (PDIS)~\citep{precup2000eligibility}
        \item Doubly Robust (DR)~\citep{jiang2016doubly, thomas2016data}
        \item Self-Normalized Trajectory-wise Importance Sampling (SNTIS)~\citep{kallus2019intrinsically}
        \item Self-Normalized Per-Decision Importance Sampling (SNPDIS)~\citep{kallus2019intrinsically}
        \item Self-Normalized Doubly Robust (SNDR)~\citep{kallus2019intrinsically}
    \end{itemize}
    \item \textbf{State Marginal Estimators}:
    \begin{itemize}
        \item (abstract base implementation)
        \item State Marginal Direct Method (SM-DM)~\citep{uehara2020minimax}
        \item State Marginal Importance Sampling (SM-IS)~\citep{liu2018breaking, uehara2020minimax, yuan2021sope}
        \item State Marginal Doubly Robust (SM-DR)~\citep{liu2018breaking, uehara2020minimax, yuan2021sope}
        \item State Marginal Self-Normalized Importance Sampling (SM-SNIS)~\citep{liu2018breaking, uehara2020minimax, yuan2021sope}
        \item State Marginal Self-Normalized Doubly Robust (SM-SNDR)~\citep{liu2018breaking, uehara2020minimax, yuan2021sope}
    \end{itemize}
    \item \textbf{State-Action Marginal Estimators}:
    \begin{itemize}
        \item (abstract base implementation)
        \item State-Action Marginal Importance Sampling (SAM-IS)~\citep{uehara2020minimax, yuan2021sope}
        \item State-Action Marginal Doubly Robust (SAM-DR)~\citep{uehara2020minimax, yuan2021sope}
        \item State-Action Marginal Self-Normalized Importance Sampling (SAM-SNIS)~\citep{uehara2020minimax, yuan2021sope}
        \item State-Action Marginal Self-Normalized Doubly Robust (SAM-SNDR)~\citep{uehara2020minimax, yuan2021sope}
    \end{itemize}
    \item \textbf{Double Reinforcement Learning}:
    \begin{itemize}
        \item Double Reinforcement Learning~\citep{kallus2020double}
    \end{itemize}
    \item \textbf{Weight and Value Learning Methods}:
    \begin{itemize}
        \item Augmented Lagrangian Method (ALM/DICE)
        \begin{itemize}
            \item BestDICE~\citep{yang2020off}
            \item GradientDICE~\citep{zhang2020gradientdice}
            \item GenDICE~\citep{zhang2020gendice}
            \item AlgaeDICE~\citep{nachum2019algaedice}
            \item DualDICE~\citep{nachum2019dualdice}
            \item MQL/MWL~\citep{uehara2020minimax}
        \end{itemize}
        \item Minimax Q-Learning and Weight Learning (MQL/MWL)~\citep{uehara2020minimax}
    \end{itemize}
    \item \textbf{High Confidence OPE}
    \begin{itemize}
        \item Bootstrap~\citep{hanna2017bootstrapping,thomas2015confidence}
        \item Hoeffding~\citep{thomas2015high}
        \item (Empirical) Bernstein~\citep{thomas2015confidence,thomas2015high}
        \item Student T-test~\citep{thomas2015confidence}
    \end{itemize}
\end{itemize}
\end{tcolorbox}

\begin{tcolorbox}[colback=red!12!white, boxrule = 0.5pt, colframe = white,]
\large{\underline{\textbf{Cumulative Distribution OPE (CD-OPE)}}}
\normalsize
\begin{itemize}
    \item \textbf{Estimators}:
    \begin{itemize}
        \item (abstract base)
        \item Direct Method (DM)~\citep{huang2021off}
        \item Trajectory-wise Importance Sampling (TIS)~\citep{chandak2021universal,huang2021off}
        \item Trajectory-wise Doubly Robust (TDR)~\citep{chandak2021universal,huang2021off}
        \item Self-Normalized Trajectory-wise Importance Sampling (SNTIS)~\citep{chandak2021universal,huang2021off}
        \item Self-Normalized Trajectory-wise Doubly Robust (SNDR)~\citep{huang2021off}
    \end{itemize}
    \item \textbf{Metrics of Interest}:
    \begin{itemize}
        \item Cumulative Distribution Function (CDF)
        \item Mean (i.e., policy value)
        \item Variance
        \item Conditional Value at Risk (CVaR)
        \item Interquartile Range
    \end{itemize}
\end{itemize}
\end{tcolorbox}

\begin{tcolorbox}[colback=blue!12!white, coltitle=black, boxrule = 0.5pt, colframe = white, colbacktitle=blue!12!white]
\large{\underline{\textbf{Evaluation-of-OPE Metrics}}}
\normalsize
\begin{itemize}
    \item \textbf{Conventional Metrics}:
    \begin{itemize}
        \item Mean Squared Error (MSE)~\citep{uehara2022review,voloshin2019empirical}
        \item Spearman's Rank Correlation Coefficient~\citep{fu2021benchmarks,paine2020hyperparameter}
        \item Regret~\citep{doroudi2017importance}
        \item Type I and Type II Error Rates
    \end{itemize}
    \item \textbf{Top-$k$ Risk-Return Tradeoff (including SharpeRatio@k~\citep{kiyohara2023towards})}
    \begin{itemize}
        \item \{Best / Worst / Mean / Std\} of \{policy value / CVaR / lower quartile\} among top-$k$ deployment policies
        \item Safety violation rate
        \item SharpeRatio~\citep{kiyohara2023towards,sharpe1998sharpe}
    \end{itemize}
\end{itemize}
\end{tcolorbox}

\section{Key Feature 1: Cumulative distribution OPE} \label{sec:cumulative_distribution_ope}

As introduced in Section~\ref{sec:implemented_ope}, SCOPE-RL implements cumulative distribution OPE (CD-OPE)~\citep{chandak2021universal,huang2021off,huang2022off}, which aims to estimate the full distribution of policy performance. To show the benefit of CD-OPE methods, we discuss the difference between (standard) OPE and CD-OPE in the following.

\paragraph{Preliminaries.}
We consider a general RL setup, formalized by a Markov Decision Process (MDP) defined by the tuple $\langle \mathcal{S}, \mathcal{A}, \mathcal{T}, P_r, \gamma \rangle$. Here, $\mathcal{S}$ represents the state space and $\mathcal{A}$ denotes the action space, which can either be discrete or continuous. Let $\mathcal{T}: \mathcal{S} \times \mathcal{A} \rightarrow \mathcal{P}(\mathcal{S})$ be the state transition probability, where $\mathcal{T}(s' | s,a)$ is the probability of observing state $s'$ after taking action $a$ in state $s$. $P_r: \mathcal{S} \times \mathcal{A} \times \mathbb{R} \rightarrow [0,1]$ represents the probability distribution of the immediate reward, and $R(s,a) := \mathbb{E}_{r \sim P_r (r | s, a)}[r]$ is the expected immediate reward when taking action $a$ in state $s$. $\pi: \mathcal{S} \rightarrow \mathcal{P}(\mathcal{A})$ denotes a \textit{policy}, where $\pi(a| s)$ is the probability of taking action $a$ in state $s$.

\paragraph{Off-Policy Evaluation (OPE).} In OPE, we are given a logged dataset collected by some \textit{behavior} policy $\pi_b$ as follows.
\begin{align*}
    \calD := \{ (s_t, a_t, s_{t+1}, r_t) \}_{t=0}^{T-1} \sim p(s_0) \prod_{t=0}^{T-1} \pi_b(a_t | s_t) \mathcal{T}(s_{t+1} | s_t, a_t) P_r (r_t | s_t, a_t).
\end{align*}
Using only the fixed logged dataset $\calD$, (standard) OPE aims to evaluate the expected reward under an evaluation (new) policy, called the \textit{policy value}. More rigorously, the policy value is defined as the expected trajectory-wise reward obtained by deploying an \textit{evaluation} policy $\pi$:
\begin{align*}
    J(\pi) := \mathbb{E}_{\tau \sim p_{\pi}(\tau)} \left[ \sum_{t=0}^{T-1} \gamma^t r_t \right],
\end{align*}
where $\gamma \in (0,1]$ is a discount factor and $p_{\pi}(\tau) = p(s_0) \prod_{t=0}^{T-1} \pi(a_t | s_t) \mathcal{T}(s_{t+1} | s_t, a_t) P_r (r_t | s_t, a_t)$ is the probability of observing a trajectory under evaluation policy $\pi$. 

While the typical definition of policy value effectively compares policies based on their expected performance, in practice, particularly in safety-critical scenarios, understanding the entire performance distribution of a policy is often more crucial and useful. For instance, in recommender systems, the aim is to consistently deliver good-quality recommendations rather than occasionally offering outstanding products while at other times significantly diminishing user satisfaction with poor choices. Similarly, in the context of self-driving cars, it is imperative to avoid catastrophic accidents, even if the probability of such events is extremely low (like less than 0.1\%). In these situations, CD-OPE proves especially valuable for assessing the performance of policies in worst-case scenarios.

\paragraph{Cumulative Distribution Off-Policy Evaluation (CD-OPE).} In contrast to the traditional approach of OPE that focuses on the point-wise estimation of expected policy performance, CD-OPE seeks to estimate the entire performance distribution of a policy. Specifically, CD-OPE focuses on estimating the CDF of policy performance, providing a more comprehensive perspective on potential consequences~\citep{chandak2021universal, huang2021off, huang2022off}:
\begin{align*}
    F(m, \pi) := \mathbb{E} \left[ \mathbb{I} \left \{ \sum_{t=0}^{T-1} \gamma^t r_t \leq m \right \} \mid \pi \right].
\end{align*}
Based on the CDF ($F(\cdot)$), we can derive various risk functions on the policy performance as follows.

\begin{enumerate}
    \item Mean: $\mu(F) := \int_{G} G \, \mathrm{d}F(G)$
    \item Variance: $\sigma^2(F) := \int_{G} (G - \mu(F))^2 \, \mathrm{d}F(G)$
    \item $\alpha$-quartile: $Q^{\alpha}(F) := \min \{ G \mid F(G) \leq \alpha \}$
    \item Conditional Value at Risk (CVaR): $\int_{G} G \, \mathbb{I}\{ G \leq Q^{\alpha}(F) \} \, \mathrm{d}F(G)$
\end{enumerate}
where we define $G := \sum_{t=0}^{T-1} \gamma^t r_t$ as the cumulative reward for a trajectory. The term $\mathrm{d}F(G) := \mathrm{lim}_{\Delta \rightarrow 0} F(G) - F(G- \Delta)$ represents the differential of the cumulative distribution function at $G$. The $\alpha$-quartile refers to the performance range extending from the lowest $100 \times \alpha \, \%$ to the highest $100 \times (1 - \alpha) \, \%$ of observations. Conditional Value at Risk (CVaR) is calculated as the average of the lowest $100 \times \alpha \, \%$ of these observations. These functions offer a more detailed analysis than just the expected performance, aiding practitioners in assessing the safety and robustness of a policy.

\begin{figure}[h]
  \centering
  \includegraphics[clip, width=\linewidth]{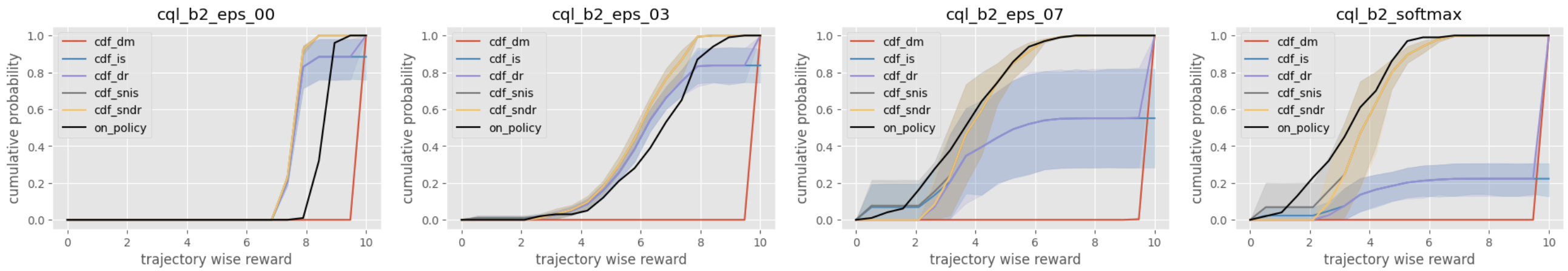}
  \vspace{-3mm}
  \caption{(Top) Example code of estimating the CDF with CD-OPE estimators implemented in SCOPE-RL. (Bottom) The output of the visualization function of the CD-OPE module.} \label{fig:cdope_example}
\end{figure}

Figure~\ref{fig:cdope_example} shows that SCOPE-RL enables implementing CD-OPE with minimal efforts.

\vspace{3mm}

\begin{lstlisting}
# import modules and estimators for cumulative distribution OPE from SCOPE-RL
from scope_rl.ope import CumulativeDistributionOPE
from scope_rl.ope.discrete import CumulativeDistributionDM as CD_DM
from scope_rl.ope.discrete import CumulativeDistributionTIS as CD_IS
from scope_rl.ope.discrete import CumulativeDistributionTDR as CD_DR
from scope_rl.ope.discrete import CumulativeDistributionSNTIS as CD_SNIS
from scope_rl.ope.discrete import CumulativeDistributionSNTDR as CD_SNDR

# initialize CumulativeDistributionOPE class
cd_ope = CumulativeDistributionOPE(
    logged_dataset=test_logged_dataset,
    ope_estimators=[CD_DM(), CD_IS(), CD_DR(), CD_SNIS(), CD_SNDR()],
)

# estimate and visualize the cumulative distribution function
cdf_dict = cd_ope.estimate_cumulative_distribution_function(input_dict)
cd_ope.visualize_cumulative_distribution_function(input_dict)
\end{lstlisting}
\vspace{-5mm}

\section{Key Feature 2: Comprehensive evaluation-of-OPE metrics} \label{sec:assessments}

\begin{figure}
  \centering
  \includegraphics[clip, width=0.70\linewidth]{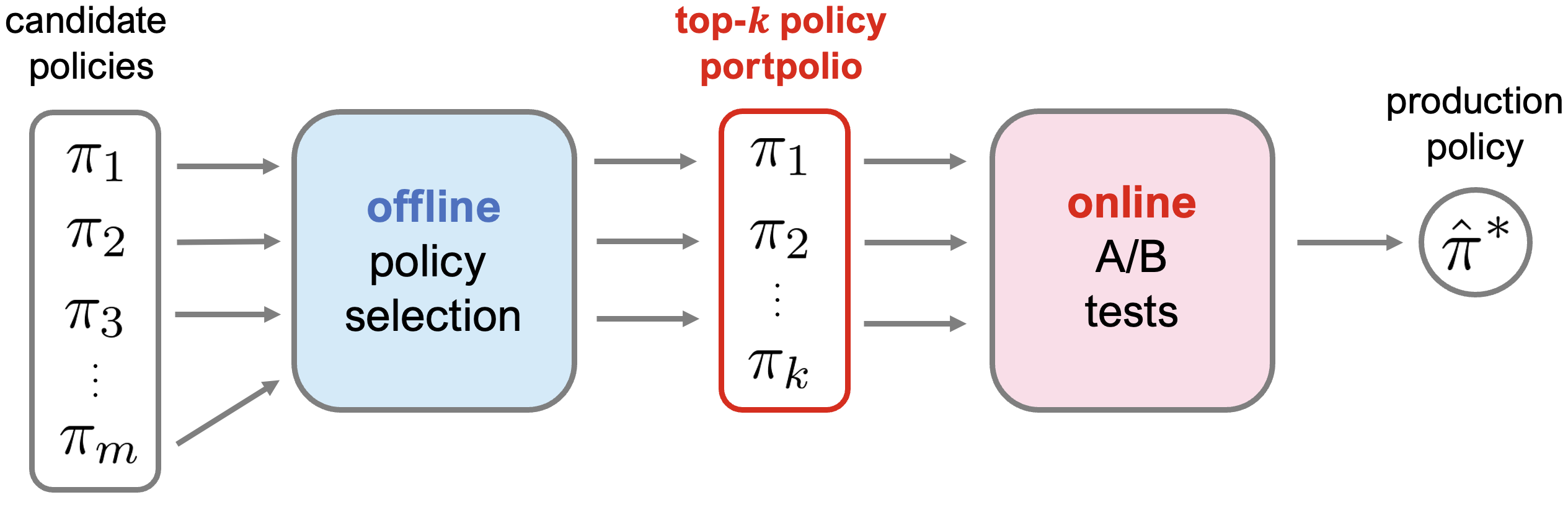}
  \caption{Practical workflow of policy evaluation and selection involves OPE as a screening process where an OPE estimator ($\hat{J}$) chooses top-$k$ (shortlisted) candidate policies that are to be tested in online A/B tests, where $k$ is a pre-defined online evaluation budget. A policy that is identified as the best policy based on the online evaluation process will be chosen as the production policy ($\hat{\pi}^*$). (Credit of the figure and description: \citep{kiyohara2023towards})} \label{fig:practical_workflow}
\end{figure}

Another distinctive feature of SCOPE-RL is to enable risk-return assessments of the downstream policy selection tasks (known as off-policy selection or OPS). 

\paragraph{Background.}
While OPE is a valuable tool for estimating the performance of new policies using offline logged data, it can sometimes yield inaccurate estimations due to bias and variance issues. Consequently, in real-world applications, it is imprudent to rely exclusively on OPE results for selecting a production policy. Instead, a combination of OPE results and online A/B testing is often employed for more comprehensive policy evaluation and selection~\citep{kiyohara2023towards,kurenkov2022showing}. Typically, the practical workflow starts by using OPE results to eliminate underperforming policies. Subsequently, A/B tests are conducted on the remaining top-$k$ policies to determine the most effective one through a more dependable online evaluation, as depicted in Figure~\ref{fig:practical_workflow}.

 \paragraph{Evaluation of OPE.}
 To evaluate and compare the effectiveness of OPE estimators, the following accuracy metrics are often used:

 \begin{itemize}
     \item \textbf{Mean Squared Error (MSE)}~\citep{uehara2022review, voloshin2019empirical}: This metric measures the estimation accuracy of estimator $\hat{J}$ among a set of policies $\Pi$ as $(1 / |\Pi|) \sum_{\pi \in \Pi} \mE_{\calD}[(\hat{J}(\pi; \mathcal{D}) - J(\pi))^2]$.
     \item \textbf{Rank Correlation (RankCorr)}~\citep{fu2021benchmarks,paine2020hyperparameter}: This metric measures how well the ranking of candidate policies is preserved in the OPE results and is defined as the spearman's rank correlation between $\{J(\pi)\}_{\pi\in \Pi}$ and $\{\hat{J}(\pi; \calD)\}_{\pi\in \Pi}$.
     \item \textbf{Regret@$k$}~\citep{doroudi2017importance}: This metric measures how well the best policy among the top-$k$ candidate policies selected by an estimator performs. In particular, Regret@1 measures the performance difference between the true best policy $\pi^{\ast}$ and the best policy estimated by the estimator as $J(\pi^{\ast}) - J(\hat{\pi}^{\ast})$ where $\hat{\pi}^{\ast} := {\arg\max}_{\pi \in \Pi} \hat{J}(\pi; \mathcal{D})$.
 \end{itemize}

In the aforementioned metrics, MSE evaluates the accuracy of OPE estimation, while the latter two metrics focus on the accuracy of downstream policy selection. By integrating these metrics, we can determine how effectively an OPE estimator selects a near-optimal policy based solely on OPE results. However, a significant limitation of this conventional approach for evaluation-of-OPE is that it fails to consider potential risks encountered during online A/B tests, especially in more practical two-stage selection processes that include online A/B testing as a final process~\citep{kiyohara2023towards}.

To remedy this issue, SCOPE-RL offers evaluation-of-OPE metrics that evaluate the risk-return tradeoff in selecting the top-$k$ policies. Our fundamental approach involves treating the set of top-$k$ candidate policies chosen by an OPE estimator as its \textit{policy portfolio}. Subsequently, we evaluate the risk, return, and efficiency of an estimator by reporting the following statistics of the top-$k$ policy portfolio:

\begin{itemize}
    \item \textbf{best@$\bm{k}$} (Return; higher is better): This metric represents the value of the highest-performing policy among the top-$k$ policies selected by an estimator. It indicates the effectiveness of the production policy chosen through top-$k$ A/B tests post-deployment.
    \item \textbf{worst@$\bm{k}$}, \textbf{mean@$\bm{k}$} (Risk; higher is better): These metrics reveal the worst and average performance among the top-$k$ policies selected by an estimator. They provide insight into how well the policies tested in A/B tests perform on average and in the worst-case scenario.
    \item \textbf{std@$\bm{k}$} (Risk; lower is better): This metric calculates the standard deviation of policy values among the top-$k$ policies chosen by an estimator. It indicates the likelihood of erroneously deploying poorly performing policies.
    \item \textbf{safety violation rate@$\bm{k}$} (Risk; lower is better): This metric quantifies the probability of policies deployed in online A/B tests violating predefined safety requirements.
    \item \textbf{SharpeRatio@k} (Efficiency; higher is better): This metric evaluates the return (best@$k$) relative to a risk-free baseline ($J(\pi_b)$), considering the risk (std@$k$) in its denominator. This provides a measure of efficiency balancing risk and return.
    \begin{align*}
        \text{SharpeRatio@}k (\hat{J}) := \frac{\text{best@}k (\hat{J}) - J(\pi_b)}{\text{std@}k(\hat{J})}.
    \end{align*}
    By comparing the SharpeRatio metric, we can identify an OPE estimator that is capable of deploying policies which not only enhance performance over the baseline but also minimize risks. Note that this metric is the main proposal of our sister paper \citep{kiyohara2023towards}.
\end{itemize}

\begin{figure}[h]
  \centering
  \includegraphics[clip, width=\linewidth]{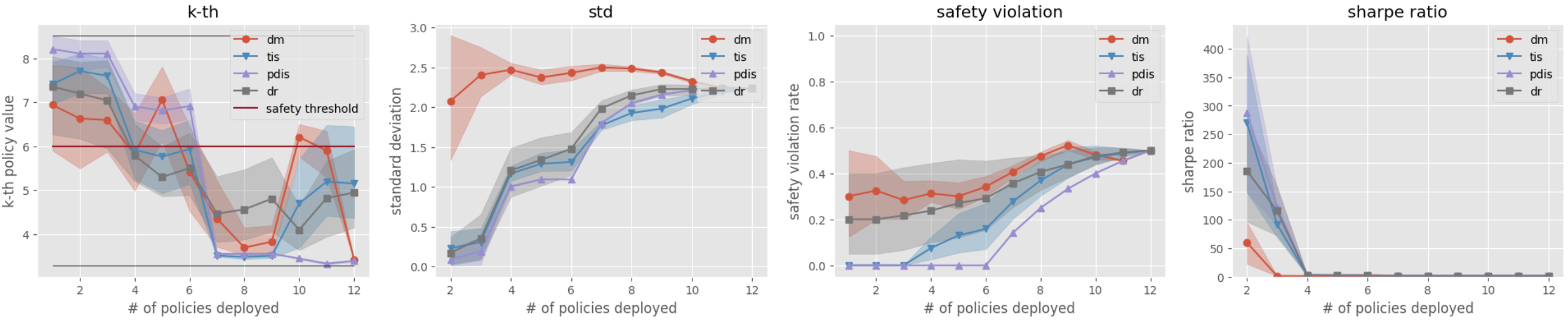}  
  \vspace{-3mm}
  \caption{(Top) Example code to perform evaluation-of-OPE with SharpRatio@k and other statistics of top-$k$ policy portfolio using SCOPE-RL. (Bottom) Visualizing the evaluation-of-OPE results.} \label{fig:topk_example}
\end{figure}

Using the SCOPE-RL package, we can also evaluate how the risk-return tradeoff metrics change with varying online evaluation budgets ($k$) in online A/B tests (See examples in Figure~\ref{fig:topk_example}).

\vspace{3mm}

\begin{lstlisting}
from scope_rl.ope import OffPolicySelection

# initialize the OPS class with OPE instances
ops = OffPolicySelection(
    ope=ope,
    cumulative_distribution_ope=cd_ope,
)

# visualize the top k deployment result
ops.visualize_topk_policy_value_selected_by_standard_ope(
    input_dict=input_dict,
    compared_estimators=["dm", "tis", "pdis", "dr"],
    metrics=["best", "worst", "std", "sharpe_ratio"],
    relative_safety_criteria=1.0,
)
\end{lstlisting}
\vspace{-5mm}

\section{User-friendly APIs, visualization tools, and documentation} \label{sec:api}

SCOPE-RL's user-friendly APIs and comprehensive support for implementation are among its key attributes. As demonstrated in Figures~\ref{fig:cdope_example} and \ref{fig:topk_example} and further detailed in Appendix~\ref{app:example}, SCOPE-RL allows for the management of the entire offline RL-to-OPE process with just a few lines of code. Additionally, visualizing results is straightforward and offers valuable insights for comparing various policies and OPE estimators.

Customizing modules in SCOPE-RL is also streamlined, thanks to our comprehensive support resources. These include API references\footnote{\textcolor{dkpink}{\href{https://scope-rl.readthedocs.io/en/latest/documentation/scope_rl_api.html}{https://scope-rl.readthedocs.io/en/latest/documentation/scope\_rl\_api.html}}}, detailed usage guides\footnote{\textcolor{dkpink}{\href{https://scope-rl.readthedocs.io/en/latest/documentation/examples/index.html}{https://scope-rl.readthedocs.io/en/latest/documentation/examples/index.html}}}, and quickstart examples\footnote{\textcolor{dkpink}{\href{https://github.com/hakuhodo-technologies/scope-rl/tree/main/examples/quickstart}{https://github.com/hakuhodo-technologies/scope-rl/tree/main/examples/quickstart}}}, all designed to provide clear instructions for various implementation options. This enables users to effortlessly test their own OPE estimators in different environmental settings or with real-world datasets. We believe SCOPE-RL greatly facilitates rapid benchmarking and practical application of offline RL and OPE in both research and practice.

\section{Summary and Future Work}

This paper introduces SCOPE-RL, a Python package tailored for offline RL and OPE, with a special emphasis on OPE modules. SCOPE-RL pioneers the implementation of cutting-edge cumulative distribution OPE (CD-OPE) and evaluation-of-OPE metrics through risk-return tradeoffs. Additionally, our extensive, well-structured documentation and intuitive APIs aid researchers and practitioners in efficiently implementing offline RL and OPE procedures.

For future developments, we aim to enhance SCOPE-RL further. Potential updates include integrating more advanced CD-OPE estimators~\citep{huang2022off,wu2023distributional,xu2022quantile}, estimators for partially observable settings~\citep{uehara2022future}, and estimator selection methods for OPE~\citep{lee2022oracle,su2020adaptive,udagawa2023policy,xie2021batch,zhang2021towards}. Adding tutorials on OPE to our documentation could also be valuable, helping users grasp OPE concepts more clearly. We also welcome and encourage pull requests, suggestions, and feedback from the user community.

Lastly, SCOPE-RL draws inspiration from \textit{OpenBanditPipeline}~\citep{saito2021open}, which has been successful in facilitating flexible OPE experiments in contextual bandits~\citep{metelli2021subgaussian,udagawa2023policy,saito2022off,saito2023off} and slate bandits~\citep{kiyohara2022doubly, kiyohara2023off, kiyohara2024off}. We believe that SCOPE-RL will similarly become a valuable tool for quick prototyping and benchmarking in the OPE of RL policies, paralleling OBP's role in non-RL contexts.

\begin{ack}
We would like to thank Koji Kawamura and Mariko Takeuchi for providing valuable feedback on the earlier version of SCOPE-RL. We would also like to thank Daniel Cao and Romain Deffayet for their helpful comments on the manuscript.
\end{ack}

\bibliographystyle{plainnat}


\clearpage

\appendix

\section{Details of implemented OPE estimators and assessment metrics} \label{app:implementation}

Here, we provide the definition and properties of OPE estimators and assessment metrics implemented in SCOPE-RL, which are listed in Section~\ref{sec:implemented_ope}.

\subsection{Standard Off-Policy Evaluation} \label{app:standard_ope}
As described in the main text, the goal of OPE in RL is to estimate the expected trajectory-wise reward under an evaluation policy $\pi$ using only the logged data $\calD$ collected by a behavior policy $\pi_b$:
\begin{align*}
    J(\pi) := \mathbb{E}_{\tau \sim p_{\pi}(\tau)} \left[ \sum_{t=0}^{T-1} \gamma^t r_t \right].
\end{align*}

\paragraph{\textbf{Direct Method (DM)}}
DM is a model-based approach, which uses the initial state value estimated by Fitted Q Evaluation (FQE)~\citep{le2019batch}.\footnote{SCOPE-RL uses the implementation of FQE provided by d3rlpy~\citep{seno2021d3rlpy}.}
It first learns the Q-function from the logged data via temporal-difference (TD) learning and then utilizes the estimated Q-function for OPE as follows.
\begin{align*}
    \hat{J}_{\mathrm{DM}} (\pi; \mathcal{D}) := \frac{1}{n} \sum_{i=1}^n \sum_{a \in \mathcal{A}} \pi(a | s_{0}^{(i)}) \hat{Q}(s_{0}^{(i)}, a) = \frac{1}{n} \sum_{i=1}^n \hat{V}(s_{0}^{(i)}),
\end{align*}
where $\hat{Q}(s_t, a_t)$ is an estimated state-action value and $\hat{V}(s_t)$ is the estimated state value. DM has lower variance compared to other estimators, but can produce large bias caused by approximation errors of the Q-function~\citep{jiang2016doubly, thomas2016data}.

\paragraph{\textbf{Trajectory-wise Importance Sampling (TIS)}}
TIS is a model-free approach, which uses the importance sampling technique to correct the distribution shift between $\pi$ and $\pi_b$ as follows~\citep{precup2000eligibility}.
\begin{align*}
    \hat{J}_{\mathrm{TIS}} (\pi; \mathcal{D}) := \frac{1}{n} \sum_{i=1}^n \sum_{t=0}^{T-1} \gamma^t w_{1:T-1}^{(i)} r_t^{(i)},
\end{align*}
where $w_{0:T-1} := \prod_{t=0}^{T-1} (\pi(a_t | s_t) / \pi_b(a_t | s_t))$ is the (trajectory-wise) importance weight.
TIS enables unbiased estimation of the policy value. However, particularly when the trajectory length $T$ and the action space $\calA$ is large, TIS suffers from high variance due to trajectory-wise importance weighting~\citep{jiang2016doubly, thomas2016data}.

\paragraph{\textbf{Per-Decision Importance Sampling (PDIS)}}
PDIS leverages the sequential nature of the MDP to reduce the variance of TIS.
Specifically, since $s_t$ only depends on the states and actions observed previously (i.e., $s_0, \ldots, s_{t-1}$ and $a_0, \ldots, a_{t-1}$) and is independent of those observed in future time steps (i.e., $s_{t+1}, \ldots, s_{T}$ and $a_{t+1}, \ldots, a_{T}$),
PDIS considers only the importance weights related to past interactions for each time step as follows~\citep{precup2000eligibility}.
\begin{align*}
    \hat{J}_{\mathrm{PDIS}} (\pi; \mathcal{D}) := \frac{1}{n} \sum_{i=1}^n \sum_{t=0}^{T-1} \gamma^t w_{0:t}^{(i)} r_t^{(i)},
\end{align*}
where $w_{0:t} := \prod_{t'=0}^t (\pi(a_{t'} \mid s_{t'}) / \pi_b(a_{t'} \mid s_{t'}))$ represents the importance weight with respect to the previous action choices for time step $t$. PDIS retains its unbiased nature while reducing the variance of TIS. However, it is well-known that PDIS can still suffer from high variance when $T$ is large~\citep{jiang2016doubly, thomas2016data}.

\paragraph{\textbf{Doubly Robust (DR)}}
DR is a hybrid of model-based estimation and importance sampling~\citep{dudik2011doubly}.
It introduces $\hat{Q}$ as a baseline estimation in the recursive form of PDIS and applies importance weighting only to its residual~\citep{jiang2016doubly, thomas2016data}. 
\begin{align*}
    \hat{J}_{\mathrm{DR}} (\pi; \mathcal{D})
    := \frac{1}{n} \sum_{i=1}^n \sum_{t=0}^{T-1} \gamma^t \left(w_{0:t}^{(i)} (r_t^{(i)} - \hat{Q}(s_t^{(i)}, a_t^{(i)})) + w_{0:t-1}^{(i)} \sum_{a \in \mathcal{A}} \pi(a | s_t^{(i)}) \hat{Q}(s_t^{(i)}, a) \right),
\end{align*}
As $\hat{Q}$ works as a control variate, DR is unbiased and at the same time reduces the variance of TIS when $\hat{Q}(\cdot)$ is reasonably accurate. However, it can still have high variance when the trajectory length $T$~\citep{fu2021benchmarks} or the action space $|\calA|$~\citep{saito2022off} is large.

\paragraph{\textbf{Self-Normalized estimators}}
Self-normalized estimators aim to reduce the scale of the importance weight for the variance reduction purpose~\citep{swaminathan2015self}. Specifically, the self-normalized versions of PDIS and DR are defined as follows.
\begin{align*}
    & \hat{J}_{\mathrm{SNPDIS}} (\pi; \mathcal{D}) 
    := \sum_{i=1}^n \sum_{t=0}^{T-1} \gamma^t \frac{w_{0:t}^{(i)}}{\sum_{i'=1}^n w_{0:t}^{(i')}} r_t^{(i)}, \\
    & \hat{J}_{\mathrm{SNDR}} (\pi; \mathcal{D}) \\
    &:= \sum_{i=1}^n \sum_{t=0}^{T-1} \gamma^t \left(\frac{w_{0:t}^{(i)}}{\sum_{i'=1}^n w_{0:t}^{(i')}} (r_t^{(i)} - \hat{Q}(s_t^{(i)}, a_t^{(i)})) + \frac{w_{0:t-1}^{(i)}}{\sum_{i'=1}^n w_{0:t-1}^{(i')}} \sum_{a \in \mathcal{A}} \pi(a | s_t^{(i)}) \hat{Q}(s_t^{(i)}, a) \right),
\end{align*}
In more general, self-normalized estimators substitute importance weight $w_{\ast}$ as $\tilde{w}_{\ast} := w_{\ast} / (\sum_{i=1}^n w_{\ast})$, where $\tilde{w}_{\ast}$ is called the self-normalized importance weight. While self-normalized estimators no longer ensures unbiasedness, they basically remain consistent. Moreover, self-normalized estimators have the variance bounded by $r_{max}^2$, which is much smaller than the variance of the original estimators~\citep{kallus2019intrinsically}.

\paragraph{\textbf{Marginalized Importance Sampling estimators}}
When the trajectory length ($T$) is large, the variance of PDIS and DR can be very high. This issue is often referred to as the curse of horizon in OPE. To alleviate this variance issue of the estimators that rely on importance weights with respect to the policies, several estimators utilize state marginal or state-action marginal importance weights, which are defined as follows~\citep{liu2018breaking, uehara2020minimax}:
\begin{align*}
    \rho(s, a) := d^{\pi}(s, a) / d^{\pi_b}(s, a), \quad \quad \quad \quad \quad \rho(s) := d^{\pi}(s) / d^{\pi_b}(s)
\end{align*}
where $d^{\pi}(s, a)$ and $d^{\pi}(s)$ is the marginal visitation probability of the policy $\pi$ on $(s, a)$ or $s$, respectively. 
The use of marginal importance weights is particularly beneficial when policy visits the same or similar states among different trajectories or different timestep.
(e.g., when the state transition is something like $\cdots \rightarrow s_1 \rightarrow s_2 \rightarrow s_1 \rightarrow s_2 \rightarrow \cdots$ or when the trajectories always visit some particular state as $\cdots \rightarrow s_{*} \rightarrow s_{1} \rightarrow s_{*} \rightarrow \cdots$).
Then, State-Action Marginal Importance Sampling (SMIS) and State Marginal Doubly Robust (SMDR) are defined as follows.
\begin{align*}
    &\hat{J}_{\mathrm{SAM-IS}} (\pi; \mathcal{D}) 
    := \frac{1}{n} \sum_{i=1}^n \sum_{t=0}^{T-1} \gamma^t \rho(s_t^{(i)}, a_t^{(i)}) r_t^{(i)}, \\
    & \hat{J}_{\mathrm{SAM-DR}} (\pi; \mathcal{D}) \\
    &:= \frac{1}{n} \sum_{i=1}^n \sum_{a \in \mathcal{A}} \pi(a | s_0^{(i)}) \hat{Q}(s_0^{(i)}, a) \\
    & \quad \quad + \frac{1}{n} \sum_{i=1}^n \sum_{t=0}^{T-1} \gamma^t \rho(s_t^{(i)}, a_t^{(i)}) \left(r_t^{(i)} + \gamma \sum_{a \in \mathcal{A}} \pi(a | s_t^{(i)}) \hat{Q}(s_{t+1}^{(i)}, a) - \hat{Q}(s_t^{(i)}, a_t^{(i)}) \right),
\end{align*}

Similarly, State-Marginal Importance Sampling (SMIS) and State Action-Marginal Doubly Robust (SAMDR) are defined as follows.

\begin{align*}
    & \hat{J}_{\mathrm{SM-IS}} (\pi; \mathcal{D}) 
    := \frac{1}{n} \sum_{i=1}^n \sum_{t=0}^{T-1} \gamma^t \rho(s_t^{(i)}) w_t(s_t^{(i)}, a_t^{(i)}) r_t^{(i)}, \\
    & \hat{J}_{\mathrm{SM-DR}} (\pi; \mathcal{D}) \\
    &:= \frac{1}{n} \sum_{i=1}^n \sum_{a \in \mathcal{A}} \pi(a | s_0^{(i)}) \hat{Q}(s_0^{(i)}, a) \\
    & \quad \quad + \frac{1}{n} \sum_{i=1}^n \sum_{t=0}^{T-1} \gamma^t \rho(s_t^{(i)}) w_t(s_t^{(i)}, a_t^{(i)}) \left(r_t^{(i)} + \gamma \sum_{a \in \mathcal{A}} \pi(a | s_t^{(i)}) \hat{Q}(s_{t+1}^{(i)}, a) - \hat{Q}(s_t^{(i)}, a_t^{(i)}) \right),
\end{align*}

\begin{table}
  \caption{Correspondence between the hyperparameter setups of Augmented Lagrangian Method (ALM)~\citep{yang2020off} and other weight learning methods.}
  \label{tab:dice}
  \centering
  \scalebox{0.875}{
  \begin{tabular}{c|cccc}
    \toprule
    hyperparameters & $\alpha_w \in [0, \infty) $ & $\alpha_Q \in [0, \infty)$ & $\alpha_r \in \{0, 1\}$ & $\lambda \in (-\infty, \infty)$ \\
    \midrule
    BestDICE~\citep{yang2020off}  &  1 & 0 & 1 & (optimize) \\
    DualDICE~\citep{nachum2019dualdice} & 0 & 1 & 0 & 0 \\
    GenDICE~\citep{zhang2020gendice} & 0 & 1 & 0 & (optimize) \\
    GradientDICE~\citep{zhang2020gradientdice} & 0 & 1 & 0 & (optimize) \\
    AlgaeDICE~\citep{nachum2019algaedice} & 1 & 0 & 1 & 0 \\
    MQL/MWL~\citep{uehara2020minimax} & 0 & 0 & 0 & 0 \\
    \bottomrule
  \end{tabular}
  }
  \vskip 0.1in
  \raggedright
  \fontsize{9pt}{9pt}\selectfont \textit{Note}:
  $\alpha_w$ and $\alpha_Q$ are hyperparameters that regularize the complexity of the weight and value functions. $\alpha_r$ is the scaling factor of the reward. $\lambda$ is the normalization constraint that enforces $\mE_{\calD_{\pi_b}}[w(s, a)]$ to be 1. For the theoretical analysis, we refer readers to \citep{yang2020off}.
\end{table}

\textbf{\textit{How to obtain state(-action) marginal importance weight?}} \quad 
To utilize marginalized importance sampling estimators, we first need to estimate the state marginal or state-action marginal importance weight. A prevalent method for this involves leveraging the relationship between the importance weights and the state-action value function, under the assumption that the state visitation probability remains consistent across various timesteps~\citep{uehara2020minimax}.
\begin{align*}
    &\mathbb{E}_{(s, a, r, s') \sim \mathcal{D}_{\pi_b}}[w(s, a) r] \\
    &= \mathbb{E}_{(s, a, r, s') \sim \mathcal{D}_{\pi_b}}[w(s, a)(Q_{\pi}(s, a) - \gamma \mathbb{E}_{a' \sim \pi(a' | s')}[Q(s', a')])] \\
    &= (1 - \gamma) \mathbb{E}_{s_0 \sim d^{\pi}(s_0), a_0 \sim \pi(a_0 | s_0)}[Q_{\pi}(s_0, a_0)]
\end{align*}

Weight learning aims to minimize the discrepancy between the middle term and the last term of the equation provided above. This is achieved when the Q-function adversarially maximizes the difference. In particular, we use the following algorithms to estimate state marginal and state-action marginal importance weights (and the corresponding state-action value function).

\begin{itemize}
    \item \textbf{Augmented Lagrangian Method (ALM/DICE)}~\citep{yang2020off}: \\
    This method simultaneously optimize both $w(s, a)$ and $Q(s, a)$ via the following objective.
    \begin{align*}
        \max_{w \leq 0} \min_{Q, \lambda} L(w, Q, \lambda),
    \end{align*}
    \begin{align*}
        & L(w, Q, \lambda) \\
        &:= (1 - \gamma) \mathbb{E}_{s_0 \sim d(s_0), a_0 \sim \pi(s_0)} [Q(s_0, a_0)] + \lambda \\
        & \quad + \mathbb{E}_{\tau_t \sim d^{\pi_b}, a_{t+1} \sim \pi(a_{t+1} | s_{t+1})} [w(s_t, a_t) (\alpha_r r_t + \gamma Q(s_{t+1}, a_{t+1}) - Q(s_t, a_t) - \lambda)] \\
        & \quad + \alpha_Q \mathbb{E}_{(s_t, a_t) \sim d^{\pi_b}} [Q^2(s_t, a_t)] - \alpha_w \mathbb{E}_{(s_t, a_t) \sim d^{\pi_b}} [w^2(s_t, a_t)]
    \end{align*}
    where $\tau_t := (s_t, a_t, r_t, s_{t+1})$ is a data tuple in the logged data. $\alpha_w \in [0, \infty)$, $\alpha_Q \in [0, \infty)$, $\alpha_r \in \{0, 1\}$, $\lambda \in (-\infty, \infty)$ are the regularization hyperparameters.
    By setting different hyperparameters, ALM is reduced to BestDICE~\citep{yang2020off}, DualDICE~\citep{nachum2019dualdice}, GenDICE~\citep{zhang2020gendice}, GradientDICE~\citep{zhang2020gradientdice}, AlgaeDICE~\citep{nachum2019algaedice}, and MQL/MWL~\citep{uehara2020minimax}. We describe the correspondence between hyperparameter setup of ALM and other algorithms in Table~\ref{tab:dice}. 
    \item \textbf{Minimax Q-Learning and Weight Learning (MQL/MWL)}~\citep{uehara2020minimax}: \\
    This method operates under the assumption that either the value function or the weight function is expressed by a function class within a reproducing kernel Hilbert space (RKHS). It optimizes solely either the value function or the weight function. \\ \\
    In particular, when learning a weight function, MWL optimizes the function approximation using the following objective:
    \begin{align*}
        \max_w L_w^2(w, Q),
    \end{align*}
    \begin{align*}
        L_w^2(w, Q) &= \mathbb{E}_{\tau_t, \tilde{\tau}_t \sim d^{\pi_b}, a_{t+1} \sim \pi(a_{t+1} | s_{t+1}), \tilde{a}_{t+1} \sim \pi(\tilde{a}_{t+1} | \tilde{s}_{t+1})}[ \\
        & \quad \quad \quad w(s_t, a_t) w(\tilde{s}_t, \tilde{a}_t) ( K((s_t, a_t), (\tilde{s}_t, \tilde{a}_t)) + K((s_{t+1}, a_{t+1}), (\tilde{s}_{t+1}, \tilde{a}_{t+1})) \\
        & \quad \quad \quad - \gamma ( K((s_t, a_t), (\tilde{s}_{t+1}, \tilde{a}_{t+1})) + K((s_{t+1}, a_{t+1}), (\tilde{s}_t, \tilde{a}_t)) ))
        ] \\
        & \quad \quad + \gamma (1 - \gamma) \mathbb{E}_{\tau_t, \tilde{\tau}_t \sim d^{\pi_b}, a_{t+1} \sim \pi(a_{t+1} | s_{t+1}), \tilde{a}_{t+1} \sim \pi(\tilde{a}_{t+1} | \tilde{s}_{t+1}), (s_0, a_0), (\tilde{s}_0, \tilde{a}_0) \sim d^{\pi}_0}[ \\
        & \quad \quad \quad w(s_t, a_t) K((s_{t+1}, a_{t+1}), (\tilde{s}_0, \tilde{a}_0)) + w(\tilde{s}_t, \tilde{a}_t) K((\tilde{s}_{t+1}, \tilde{a}_{t+1}), (s_0, a_0))
        ] \\
        & \quad \quad - (1 - \gamma) \mathbb{E}_{(s_t, a_t), (\tilde{s}_t, \tilde{a}_t) \sim d^{\pi_b}, s_0 \sim d(s_0), \tilde{s}_0 \sim d(\tilde{s}_0), a_0 \sim \pi(a_0 | s_0), \tilde{a}_0 \sim \pi(\tilde{a}_0 | \tilde{s}_0)}[ \\
        & \quad \quad \quad w(s_t, a_t) K((s_t, a_t), (\tilde{s}_0, \tilde{a}_0)) + w(\tilde{s}_t, \tilde{a}_t) K((\tilde{s}_t, \tilde{a}_t), (s_0, a_0))
        ]
    \end{align*}
    where $\tau_t := (s_t, a_t, r_t, s_{t+1})$ is a data tuple and $K(\cdot, \cdot)$ is a kernel function.
    $(\tilde{s}_0, \tilde{a}_0) \sim d^{\pi}_0$ indicates that the initial state is sampled as $s_0 \sim d(s_0)$ and the initial action is sampled as $a_0 \sim \pi(a_0 | s_0)$. \\
    \\
    In contrast, when learning a Q-function, MQL learns from the following objective.
    \begin{align*}
        \max_Q L_Q^2(w, Q),
    \end{align*}
    \begin{align*}
        L_Q^2(w, Q)
        &= \mathbb{E}_{\tau_t, \tilde{\tau}_t \sim d^{\pi_b}, a_{t+1} \sim \pi(a_{t+1} | s_{t+1}), \tilde{a}_{t+1} \sim \pi(\tilde{a}_{t+1} | \tilde{s}_{t+1})}[ \\
        & \quad \quad \quad (r_t + \gamma Q(s_{t+1}, a_{t+1}) - Q(s_t, a_t)) \\
        & \quad \quad \quad \cdot K((s_t, a_t), (\tilde{s}_t, \tilde{a}_t)) (\tilde{r}_t + \gamma Q(\tilde{s}_{t+1}, \tilde{a}_{t+1}) - Q(\tilde{s}_t, \tilde{a}_t))] 
    \end{align*}
    where $\tau_t := (s_t, a_t, r_t, s_{t+1})$ is a data tuple and $K(\cdot, \cdot)$ is a kernel function.
\end{itemize}

\paragraph{\textbf{Double Reinforcement Learning (DRL)}}
DRL~\citep{kallus2020double} leverages marginal importance sampling in the definition of DR as follows.
\begin{align*}
    & \hat{J}_{\mathrm{DRL}} (\pi; \mathcal{D}) \\
    & := \frac{1}{n} \sum_{k=1}^K \sum_{i=1}^{n_k} \sum_{t=0}^{T-1} (\rho^j(s_{t}^{(i)}, a_{t}^{(i)}) (r_{t}^{(i)} - Q^j(s_{t}^{(i)}, a_{t}^{(i)}))
    + \rho^j(s_{t-1}^{(i)}, a_{t-1}^{(i)}) \sum_{a \in \mathcal{A}} \pi(a | s_t^{(i)}) Q^j(s_{t}^{(i)}, a)).
\end{align*}
DRL achieves the semiparametric efficiency bound with a consistent value estimator $\hat{Q}$. To alleviate the potential bias introduced in $\hat{Q}$, DRL employs the \textit{cross-fitting} technique to estimate the value function. Specifically, let $K$ represent the number of folds and $\mathcal{D}_j$ denote the $j$-th split of logged data, consisting of $n_k$ samples. The cross-fitting procedure obtains $\hat{w}^j$ and $\hat{Q}^j$ on the subset of data used for OPE, that is, $\mathcal{D} \setminus \mathcal{D}_j$.

\paragraph{\textbf{Spectrum of Off-Policy Estimators (SOPE)}}
While the state marginal or state-action marginal importance weights effectively alleviate the variance issue of per-decision importance weighting, particularly when the trajectory is long, the estimation error of marginal importance weights may introduce some bias in the estimation. To alleviate this and control the bias-variance tradeoff more flexibly, SOPE uses the following interpolated importance weights~\citep{yuan2021sope}.
\begin{align*}
    w_{\mathrm{SOPE}}(s_t, a_t) &= 
    \begin{cases}
        \prod_{t'=0}^{k-1} w_t(s_{t'}, a_{t'}) & \mathrm{if} \, t < k \\
        \rho(s_{t-k}, a_{t-k}) \prod_{t'=t-k+1}^{t} w_t(s_{t'}, a_{t'}) & \mathrm{otherwise}
    \end{cases} \\
    w_{\mathrm{SOPE}}(s_t, a_t) &= 
    \begin{cases}
        \prod_{t'=0}^{k-1} w_t(s_{t'}, a_{t'}) & \mathrm{if} \, t < k \\
        \rho(s_{t-k}) \prod_{t'=t-k}^{t} w_t(s_{t'}, a_{t'}) & \mathrm{otherwise}
    \end{cases}
\end{align*}
where SOPE uses the per-decision importance weight $w_t(s_t, a_t) := \pi(a_t | s_t) / \pi_b(a_t | s_t)$ for the $k$ most recent timesteps.

For instance, State Action-Marginal Importance Sampling (SAMIS) and State Action-Marginal Doubly Robust (SAM-DR) are defined as follows.

\begin{align*}
    &\hat{J}_{\mathrm{SOPE-SAM-IS}} (\pi; \mathcal{D})
    := \frac{1}{n} \sum_{i=1}^n \sum_{t=0}^{k-1} \gamma^t w_{0:t}^{(i)} r_t^{(i)} 
    + \frac{1}{n} \sum_{i=1}^n \sum_{t=k}^{T-1} \gamma^t \rho(s_{t-k}^{(i)}, a_{t-k}^{(i)}) w_{t-k+1:t}^{(i)} r_t^{(i)}, \\
    &\hat{J}_{\mathrm{SOPE-SAM-DR}} (\pi; \mathcal{D}) \\
    &:= \frac{1}{n} \sum_{i=1}^n \sum_{a \in \mathcal{A}} \pi(a | s_0^{(i)}) \hat{Q}(s_0^{(i)}, a) \\
    & \quad \quad + \frac{1}{n} \sum_{i=1}^n \sum_{t=0}^{k-1} \gamma^t w_{0:t}^{(i)} \left(r_t^{(i)} + \gamma \sum_{a \in \mathcal{A}} \pi(a | s_t^{(i)}) \hat{Q}(s_{t+1}^{(i)}, a) - \hat{Q}(s_t^{(i)}, a_t^{(i)}) \right) \\
    & \quad \quad + \frac{1}{n} \sum_{i=1}^n \sum_{t=k}^{T-1} \gamma^t \rho(s_{t-k}^{(i)}, a_{t-k}^{(i)}) w_{t-k+1:t}^{(i)} \left(r_t^{(i)} + \gamma \sum_{a \in \mathcal{A}} \pi(a | s_t^{(i)}) \hat{Q}(s_{t+1}^{(i)}, a) - \hat{Q}(s_t^{(i)}, a_t^{(i)}) \right),
\end{align*}

\paragraph{\textbf{High Confidence Off-Policy Evaluation}}
To mitigate the risk of overestimating the policy value due to high variance, we sometimes aim to estimate both the confidence interval and an appropriate lower bound of the policy value. SCOPE-RL implements four methods to estimate these confidence intervals~\citep{thomas2015confidence, thomas2015high}.
\begin{enumerate}
    \item Hoeffding: $\, |\hat{J}(\pi; \mathcal{D}) - \mathbb{E}_{\mathcal{D}}[\hat{J}(\pi; \mathcal{D})]| \leq \hat{J}_{\max} \displaystyle \sqrt{\frac{\log(1 / \alpha)}{2 n}}$
    \item Empirical Bernstein: $\, |\hat{J}(\pi; \mathcal{D}) - \mathbb{E}_{\mathcal{D}}[\hat{J}(\pi; \mathcal{D})]| \leq \displaystyle \frac{7 \hat{J}_{\max} \log(2 / \alpha)}{3 (n - 1)} + \displaystyle \sqrt{\frac{2 \hat{\mathbb{V}}_{\mathcal{D}}(\hat{J}) \log(2 / \alpha)}{(n - 1)}}$
    \item Student T-test: $\, |\hat{J}(\pi; \mathcal{D}) - \mathbb{E}_{\mathcal{D}}[\hat{J}(\pi; \mathcal{D})]| \leq \displaystyle \frac{T_{\mathrm{test}}(1 - \alpha, n-1)}{\sqrt{n} / \hat{\sigma}}$
    \item Bootstrapping: $\, |\hat{J}(\pi; \mathcal{D}) - \mathbb{E}_{\mathcal{D}}[\hat{J}(\pi; \mathcal{D})]| \leq \mathrm{Bootstrap}(1 - \alpha)$
\end{enumerate}
All the above bounds hold with a probability of $1 - \alpha$. In terms of notation, we denote $\hat{\mathbb{V}}_{\mathcal{D}}(\cdot)$ as the sample variance, $T_{\mathrm{test}}(\cdot,\cdot)$ as the T-value, and $\sigma$ as the standard deviation. Among the above high confidence interval estimations, the Hoeffding and empirical Bernstein methods derive lower bounds without imposing any distribution assumption on the reward, which sometimes results in overly conservative estimations. On the other hand, the T-test is based on the assumption that each sample follows a normal distribution. Thus, when the true distribution of an estimator is highly skewed, the lower bound based T-test may not hold, but it can derive a tighter bound compared to those of the Hoeffding and empirical Bernstein methods when the assumption holds.

\paragraph{\textbf{Extension to the continuous action space}}
When the action space is continuous, the naive importance weight $w_t = \pi(a_t|s_t) / \pi_b(a_t|s_t) = \int_{a \in \mathcal{A}} (\pi(a |s_t) / \pi_b(a_t|s_t)) \mathbb{I}(a = a_t) da$ ends up rejecting almost all actions, as $\mathbb{I}(a = a_t)$ filters only the action observed in the logged data. To address this issue, continuous OPE estimators apply the kernel density estimation technique to smooth the importance weight as follows~\citep{kallus2018policy}.
\begin{align*}
    \overline{w}_t = \int_{a \in \mathcal{A}} \frac{\pi(a | s_t)}{\pi_b(a_t | s_t)} \cdot \frac{1}{h} K \left( \frac{a - a_t}{h} \right) da, 
\end{align*}
where $K(\cdot)$ denotes a kernel function and $h$ is the bandwidth hyperparameter. A large value of $h$ leads to a high-bias but low-variance estimator, while a small value of $h$ results in a low-bias but high-variance estimator. Any function that can be represented as $K(\cdot)$ and satisfies the following regularity conditions can be used as the kernel function:
\begin{enumerate}
    \item $\int xK(x) dx = 0$
    \item $\int K(x) dx = 1$
    \item $\lim _{x \rightarrow-\infty} K(x)=\lim _{x \rightarrow+\infty} K(x)=0$
    \item $K(x) \geq 0, \forall x$
\end{enumerate}
We provide the following kernel functions in SCOPE-RL.
\begin{itemize}
    \item Gaussian kernel: $K(x) = \frac{1}{\sqrt{2 \pi}} e^{-\frac{x^{2}}{2}}$
    \item Epanechnikov kernel: $K(x) = \frac{3}{4} (1 - x^2) \, (|x| \leq 1)$
    \item Triangular kernel: $K(x) = 1 - |x| \, (|x| \leq 1)$
    \item Cosine kernel: $K(x) = \frac{\pi}{4} \mathrm{cos} \left( \frac{\pi}{2} x \right) \, (|x| \leq 1)$
    \item Uniform kernel: $K(x) = \frac{1}{2} \, (|x| \leq 1)$
\end{itemize}

\subsection{Cumulative Distribution Off-Policy Evaluation} \label{app:cd_ope}
In practical situations, we often have a greater interest in risk functions such as Conditional Value at Risk (CVaR) and the interquartile range of the trajectory-wise reward under an evaluation policy, rather than the mere expectation (i.e., policy value). To derive these risk functions, Cumulative Distribution Off-Policy Evaluation (CD-OPE) first estimates the following cumulative distribution function (CDF)~\citep{chandak2021universal, huang2021off, huang2022off}.
\begin{align*}
    F(m, \pi) := \mathbb{E} \left[ \mathbb{I} \left \{ \sum_{t=0}^{T-1} \gamma^t r_t \leq m \right \} \mid \pi \right],
\end{align*}
which allows us to derive various risk functions based on $F(\cdot)$ as follows.

\begin{enumerate}
    \item Mean: $\mu(F) := \int_{G} G \, \mathrm{d}F(G)$
    \item Variance: $\sigma^2(F) := \int_{G} (G - \mu(F))^2 \, \mathrm{d}F(G)$
    \item $\alpha$-quartile: $Q^{\alpha}(F) := \min \{ G \mid F(G) \leq \alpha \}$
    \item Conditional Value at Risk (CVaR): $\int_{G} G \, \mathbb{I}\{ G \leq Q^{\alpha}(F) \} \, \mathrm{d}F(G)$
\end{enumerate}
where we let $G := \sum_{t=0}^{T-1} \gamma^t r_t$ to denote the trajectory wise reward and $\mathrm{d}F(G) := \mathrm{lim}_{\Delta \rightarrow 0} F(G) - F(G- \Delta)$. $\alpha$-quartile is the performance range from $100 \times \alpha \, \%$ to $100 \times (1 - \alpha) \, \%$. CVaR is the average among the lower $100 \times \alpha \, \%$ of the observations. These functions provide more fine-grained information about the policy performance than the expected trajectory-wise reward.

Below, we describe estimators for estimating the CDF ($F(m, \pi)$) supported by SCOPE-RL.

\paragraph{\textbf{Direct Method (DM)}}
DM adopts a model-based approach to estimate the cumulative distribution function (CDF)~\citep{huang2021off}.
\begin{align*}
    \hat{F}_{\mathrm{DM}}(m, \pi; \mathcal{D}) := \frac{1}{n} \sum_{i=1}^n \sum_{a \in \mathcal{A}} \pi(a | s_0^{(i)}) \hat{G}(m; s_0^{(i)}, a)
\end{align*}
where $\hat{F}(\cdot)$ is the estimated CDF and $\hat{G}(\cdot)$ is an estimator for $\mE\left[\sum_{t=0}^{T-1} \gamma^t r_t \mid s, a \right]$. DM is vulnerable to the approximation error and resulting bias issue, but has lower variance than other estimators, similar to the basic OPE.

\paragraph{\textbf{Trajectory-wise Importance Sampling (TIS)}}
TIS corrects the distribution shift by applying the importance sampling technique on the CDF estimation~\citep{chandak2021universal,huang2021off}.
\begin{align*}
    \hat{F}_{\mathrm{TIS}}(m, \pi; \mathcal{D}) := \frac{1}{n} \sum_{i=1}^n w_{0:T-1}^{(i)} \mathbb{I} \left \{\sum_{t=0}^{T-1} \gamma^t r_t^{(i)} \leq m \right \}
\end{align*}
TIS is unbiased but can suffer from high variance. As a consequence, $\hat{F}_{\mathrm{TIS}}(\cdot)$ sometimes becomes more than 1.0 when the variance is high. Therefore, we correct CDF as follows~\citep{huang2021off}.
\begin{align*}
    \hat{F}^{\ast}_{\mathrm{TIS}}(m, \pi; \mathcal{D}) := \min(\max_{m' \leq m} \hat{F}_{\mathrm{TIS}}(m', \pi; \mathcal{D}), 1).
\end{align*}

\paragraph{\textbf{Trajectory-wise Doubly Robust (TDR)}}
TDR combines TIS and DM as follows~\citep{huang2021off}.
\begin{align*}
    \hat{F}_{\mathrm{TDR}}(m, \pi; \mathcal{D})
    := \frac{1}{n} \sum_{i=1}^n w_{0:T-1}^{(i)} \left( \mathbb{I} \left \{\sum_{t=0}^{T-1} \gamma^t r_t^{(i)} \leq m \right \} - \hat{G}(m; s_0^{(i)}, a_0^{(i)}) \right)
    + \hat{F}_{\mathrm{DM}}(m, \pi; \mathcal{D})
\end{align*}
TDR reduces the variance of TIS while being unbiased, leveraging the model-based estimate (i.e., DM) as a control variate. Since $\hat{F}_{\mathrm{TDR}}(\cdot)$ may take a value outsize the required rage of [0, 1], we should apply the following transformation to bound $\hat{F}_{\mathrm{TDR}}(\cdot) \in [0, 1]$~\citep{huang2021off}.
\begin{align*}
    \hat{F}^{\ast}_{\mathrm{TDR}}(m, \pi; \mathcal{D}) := \mathrm{clip}(\max_{m' \leq m} \hat{F}_{\mathrm{TDR}}(m', \pi; \mathcal{D}), 0, 1).
\end{align*}
Note that this estimator is not equivalent to the (recursive) DR estimator defined by~\citep{huang2022off}. We plan to implement the recursive version in future updates of the software.

\paragraph{\textbf{Self-Normalized estimators}} We also provide self-normalized estimators for the CDF, which normalize the importance weight as $\tilde{w}_{\ast} := w_{\ast} / \sum_{i=1}^n w{\ast}$ for variance reduction. Using the self-normalized importance weights, $\hat{F}_{\mathrm{SNTIS}}(\cdot)$ never exceeds one. On the other hand, $\hat{F}_{\mathrm{SNTDR}}(\cdot)$ still requires clipping to keep $\hat{F}_{\mathrm{SNTDR}}(\cdot)$ within the range of $[0, 1]$.

\paragraph{\textbf{Plans for additional implementations}} Cumulative Distribution Off-Policy Evaluation (CD-OPE) is garnering increasing attention and has become an active area of research. While we currently provide the baseline estimators proposed by \citep{chandak2021universal,huang2021off}, we aim to continue adding advanced CD-OPE estimators in future updates. For instance, the addition of a recursive DR estimator proposed by \citep{huang2022off} would be beneficial for reducing variance caused by the trajectory-wise importance weights. Additionally, the inclusion of generative modeling-based OPE estimators~\citep{wu2023distributional,xu2022quantile} will enable more flexible control of the bias-variance tradeoff via the bandwidth parameter of the kernel function.

\subsection{Evaluation metrics for OPE and OPS}
SCOPE-RL provides both conventional evaluation protocols and risk-return tradeoff metrics to evaluate OPE and OPS methods. First, we implement the following four metrics to measure the accuracy of OPE estimators, the first three of which are the baseline conventional metrics described in the main text:
\begin{itemize}
    \item \textbf{Mean Squared Error (MSE)}~\citep{voloshin2019empirical}: 
    This metric measures the estimation accuracy of estimator $\hat{J}$ across a set of policies $\Pi$ as follows:
    \begin{align*}
        \mathrm{MSE}(\hat{J}) := \frac{1}{|\Pi|} \sum_{\pi \in \Pi} \mE_{\calD}[(\hat{J}(\pi; \mathcal{D}) - J(\pi))^2].
    \end{align*}
    \item \textbf{Rank Correlation (Rankcorr)}~\citep{fu2021benchmarks,paine2020hyperparameter}:
    This metric measures how well the ranking of the candidate policies is preserved in the OPE results. It is defined as the (expected) Spearman's rank correlation between $\{J(\pi)\}_{\pi\in \Pi}$ and $\{\hat{J}\{\pi; \calD)\}_{\pi\in \Pi}$ as 
    \begin{align*}
        \mathrm{RankCorr}(\hat{J}) := \frac{\mathrm{cov}(R_{J}(\Pi), R_{\hat{J}}(\Pi))}{\mathrm{std}(R_{J}(\Pi)) \,\mathrm{std}(R_{\hat{J}}(\Pi))} = 1 - \frac{6 \sum_{\pi \in \Pi} (R_{J}(\pi) - R_{\hat{J}}(\pi))^2 }{k (k^2 - 1)},
    \end{align*}
    where $R_{J}(\cdot)$ is the ranking of candidate policy based on $J(\cdot)$, while $R_{\hat{J}}(\cdot)$ is the ranking estimated by an OPE estimator $\hat{J}$. $\mathrm{cov}(R_{J}(\Pi), R_{\hat{J}}(\Pi))$ is the covariance between the two rankings, and $\mathrm{std}(R_{\cdot}(\Pi))$ is the standard deviation of the ranking indexes, which is constant across all possible rankings.
    \item \textbf{Regret@$\bm{k}$}~\citep{doroudi2017importance}: 
    This metric measures how well the best policy among the top-$k$ candidate policies selected by an estimator performs. It is defined as follows:
    \begin{align*}
        \text{Regret@}k(\hat{J}) := \max_{\pi \in \Pi} J(\pi) - \max_{\pi \in \Pi_k(\hat{J})} J(\pi).
    \end{align*}
    In particular, Regret@1 measures the performance difference between the true best policy $\pi^{\ast}$ and the best policy estimated by an estimator as $J(\pi^{\ast}) - J(\hat{\pi}^{\ast})$, where $\hat{\pi}^{\ast} := {\arg\max}_{\pi \in \Pi} \hat{J}(\pi; \mathcal{D})$.
    \item \textbf{Type I and Type II Error Rates}: These metrics measure how well an OPE estimator validates whether the policy performance surpasses the given safety threshold or not. Below are the definitions of the Type I and Type II error rates:
    \begin{align*}
        \text{Type I error rate }(\hat{J}) &:= \frac{\sum_{\pi \in \Pi} \mathbb{I} \, \{ \hat{J}(\pi) \geq \bar{J} \} \cap \mathbb{I} \, \{ J(\pi) < \bar{J} \} }{\sum_{\pi \in \Pi} \mathbb{I} \, \{ J(\pi) < \bar{J} \} }, \\
        \text{Type II error rate }(\hat{J}) &:= \frac{\sum_{\pi \in \Pi} \mathbb{I} \, \{ \hat{J}(\pi) < \bar{J} \} \cap \mathbb{I} \, \{ J(\pi) \geq \bar{J} \} }{\sum_{\pi \in \Pi} \mathbb{I} \, \{ J(\pi) \geq \bar{J} \} },
    \end{align*}
    where $\mathbb{I} \, \{ \cdot \}$ is the indicator function and $\bar{J}$ is a safety threshold.
\end{itemize}

In addition to the above metrics, we measure the top-$k$ deployment performance to evaluate the outcome of policy selection:
\begin{itemize}
    \item \textbf{best@$\bm{k}$} (return; the larger, the better): This metric reports the best policy performance among the selected top-$k$ policies as 
    \begin{align*}
        \text{best@}k (\hat{J}) := \max_{\pi \in \Pi_k(\hat{J})} J(\pi).
    \end{align*}
    Similar to regret@$k$, this metric measures how well an OPE estimator identifies a high-performing policy.
    \item \textbf{worst@$\bm{k}$}, \textbf{mean@$\bm{k}$} (risk; the larger, the better): These metrics report the worst and mean performance among the top-$k$ policies selected by an estimator as 
    \begin{align*}
        \text{worst@}k (\hat{J}) := \min_{\pi \in \Pi_k(\hat{J})} J(\pi), \quad \quad
        \text{mean@}k (\hat{J}) := \frac{1}{|\Pi_k(\hat{J})|} \sum_{\pi \in \Pi_k(\hat{J})} J(\pi).
    \end{align*}
    These metrics quantify how likely an OPE estimator mistakenly chooses poorly-performing policies as promising.
    \item \textbf{std@$\bm{k}$} (risk; the smaller, the better): This metric reports how the performance among top-$k$ policies deviates from each other as
    \begin{align*}
        \text{std@}k(\hat{J}) :&= \sqrt{ \frac{1}{k} \sum_{\pi \in \Pi_k(\hat{J})} \left(J(\pi) - \left( \frac{1}{k} \sum_{\pi \in \Pi_k(\hat{J})} J(\pi) \right) \right)^2 }.
    \end{align*}
    This metric also quantifies how likely an OPE estimator is to mistakenly chooses poorly-performing policies.
    \item \textbf{safety violation rate@$\bm{k}$} (risk; the smaller, the better): This metric reports the probability of deployed policies violating a pre-defined safety requirement $\bar{J}$ (such as the performance of the behavior policy) as follows.
    \begin{align*}
        \text{safety violation rate@}k(\hat{J}) :&= \frac{1}{|\Pi_k(\hat{J})|} \sum_{\pi \in \Pi_k(\hat{J})} \mathbb{I} \, \{J(\pi) < \bar{J} \}.
    \end{align*}
    \item \textbf{Sharpe ratio@$\bm{k}$} (efficiency; the larger, the better): Analogous to the original Sharpe ratio used in the field of finance~\citep{sharpe1998sharpe}, we define this metric as 
    \begin{align*}
        \text{SharpeRatio@}k (\hat{J}) := \frac{\text{best@}k (\hat{J}) - J(\pi_b)}{\text{std@}k(\hat{J})}.
    \end{align*}
    This metric measures the return (best@$k$) over the risk-free baseline ($J(\pi_b)$) while accounting for risk (std@$k$) in the denominator~\citep{kiyohara2023towards}.
\end{itemize}
These metrics can be seen as the statistics of the \textit{policy portfolio} formed by a given OPE estimator. Note that these metrics are also applicable not only to standard OPE but also to cumulative distribution OPE (e.g., our implementation can measure the top-$k$ performance metric of CVaR instead of the policy value $J(\pi)$).

\newpage

\section{Example codes and tutorials of SCOPE-RL} \label{app:example}

Here, we provide some example end-to-end codes to implement offline RL, OPE, and assessments of OPE and OPS via SCOPE-RL. For more detailed usages, please also refer to \textcolor{dkpink}{\href{https://scope-rl.readthedocs.io/en/latest/documentation/examples/index.html}{https://scope-rl.readthedocs.io/en/latest/documentation/examples/index.html}}.

\subsection{Handling a single logged dataset}
We first show the case of using a single logged dataset generated by a single behavior policy. Note that we use a sub-package of SCOPE-RL called \textit{``BasicGym''} as a simple and synthetic RL environment in the following examples.

\begin{lstlisting}[title={Code Snippet 1: Setting up an synthetic environment},captionpos=b]
# setting up basicgym
import gym
import basicgym
env = gym.make("BasicEnv-discrete-v0")
\end{lstlisting}

\subsubsection{Synthetic Data Generation}
To perform an end-to-end process of offline RL and OPE, we need a logged dataset generated by a behavior policy. 
Thus, we first train a base behavior policy using d3rlpy~\citep{seno2021d3rlpy} as follows.
\begin{lstlisting}[title={Code Snippet 2: Training a base behavior policy},captionpos=b]
# behavior policy
from d3rlpy.algos import DoubleDQNConfig
from d3rlpy.models.encoders import VectorEncoderFactory
from d3rlpy.models.q_functions import MeanQFunctionFactory
from d3rlpy.dataset import create_fifo_replay_buffer
from d3rlpy.algos import LinearDecayEpsilonGreedy

# model
ddqn = DoubleDQNConfig(
    encoder_factory=VectorEncoderFactory(hidden_units=[30, 30]),
    q_func_factory=MeanQFunctionFactory(),
).create()

# replay buffer
buffer = create_fifo_replay_buffer(
    limit=10000,
    env=env,
)

# explorers
explorer = LinearDecayEpsilonGreedy(
    start_epsilon=1.0,
    end_epsilon=0.1,
    duration=1000,
)

# online training
ddqn.fit_online(
    env,
    buffer,
    explorer=explorer,
    eval_env=env,
    n_steps=100000,
)

\end{lstlisting}

\newpage

After obtaining a base behavior policy, we make it stochastic and generate logged datasets based on it. The two independent logged datasets correspond are used for performing offline RL and OPE, respectively.

\begin{lstlisting}[title={Code Snippet 3: Generating a synthetic logged dataset},captionpos=b]
# import SCOPE-RL modules
from scope_rl.dataset import SyntheticDataset
from scope_rl.policy import EpsilonGreedyHead

# converting the behavior policy to a stochastic one
behavior_policy = EpsilonGreedyHead(
    ddqn, 
    n_actions=env.action_space.n,
    epsilon=0.3,
    name="ddqn_epsilon_0.3",
    random_state=12345,
)

# initialize the dataset class
dataset = SyntheticDataset(
    env=env,
    max_episode_steps=env.step_per_episode,
)

# generate logged data for offline RL
train_logged_dataset = dataset.obtain_episodes(
    behavior_policies=behavior_policy,
    n_trajectories=10000, 
    random_state=12345,
)
# generate logged data for OPE
test_logged_dataset = dataset.obtain_episodes(
    behavior_policies=behavior_policy,
    n_trajectories=10000, 
    random_state=12345 + 1,
)
\end{lstlisting}

\subsubsection{Offline Reinforcement Learning}
Next, we train new (and hopefully better) policies from only offline logged data. Since we use d3rlpy~\citep{seno2021d3rlpy} for this offline RL part, we first transform the logged dataset to a d3rlpy format. As shown below, SCOPE-RL provides a smooth integration with d3rlpy.

\begin{lstlisting}[title={Code Snippet 4: Dataset compatibility with d3rlpy},captionpos=b]
# import d3rlpy modules
from d3rlpy.dataset import MDPDataset

# transform offline dataset to a d3rlpy format
offlinerl_dataset = MDPDataset(
    observations=train_logged_dataset["state"],
    actions=train_logged_dataset["action"],
    rewards=train_logged_dataset["reward"],
    terminals=train_logged_dataset["done"],
)
\end{lstlisting}

\newpage

Then, we train several offline RL algorithms as follows.

\begin{lstlisting}[title={Code Snippet 5: Offline Reinforcement Learning},captionpos=b]
from d3rlpy.algos import DiscreteCQLConfig as CQLConfig

# prepare models
cql_b1 = CQLConfig(
    encoder_factory=VectorEncoderFactory(hidden_units=[30, 30]),
    q_func_factory=MeanQFunctionFactory(),
).create()
cql_b2 = CQLConfig(
    encoder_factory=VectorEncoderFactory(hidden_units=[100]),
    q_func_factory=MeanQFunctionFactory(),
).create()
cql_b3 = CQLConfig(
    encoder_factory=VectorEncoderFactory(hidden_units=[50, 10]),
    q_func_factory=MeanQFunctionFactory(),
).create()
algos = [cql_b1, cql_b2, cql_b3]

# learn base evaluation policies
for i in range(len(algos)):
    algos[i].fit(
        offlinerl_dataset,
        n_steps=10000,
    )

# make stochastic candidate policies
evaluation_policies = []
for i in range(len(algos)):
    for j, epsilon in enumerate([0.3, 0.5, 0.7]):
        eval_policy = EpsilonGreedyHead(
            base_policy=algos[i], 
            n_actions=env.action_space.n, 
            name=f"cql_b{i+1}_epsilon_{epsilon}", 
            epsilon=epsilon, 
            random_state=12345,
        )
        evaluation_policies.append(eval_policy)

\end{lstlisting}

\subsubsection{Off-Policy Evaluation}
After deriving several candidate policies, we go on to evaluate their performance (policy value) using the logged data via OPE. Here, we use the following OPE estimators implemented in SCOPE-RL.

\begin{lstlisting}[title={Code Snippet 6: Representative OPE estimators implemented in SCOPE-RL},captionpos=b]
# basic estimators
from scope_rl.ope.discrete import DirectMethod as DM
from scope_rl.ope.discrete import SelfNormalizedPDIS as SNPDIS
from scope_rl.ope.discrete import SelfNormalizedDR as SNDR
# marginal importance sampling-based estimators
from scope_rl.ope.discrete import StateMarginalSNIS as SMSNIS
from scope_rl.ope.discrete import StateMarginalSNDR as SMSNDR
from scope_rl.ope.discrete import StateActionMarginalSNIS as SAMSNIS
from scope_rl.ope.discrete import StateActionMarginalSNDR as SAMSNDR
from scope_rl.ope.discrete import DoubleReinforcementLearning as DRL

# initializing OPE estimators
ope_estimators = [
    DM(), SNPDIS(), SNDR(), SMSNIS(), SMSNDR(), SAMSNIS(), SAMSNDR(), DRL(),
]
\end{lstlisting}

Even with some advanced OPE estimators, such as marginal importance sampling-based estimators, SCOPE-RL enables OPE process in an easily implementable way as follows.

\begin{lstlisting}[title={Code Snippet 7: \textbf{Basic Off-Policy Evaluation}},captionpos=b]
# import ope modules from SCOPE-RL
from scope_rl.ope import CreateOPEInput
from scope_rl.ope import OffPolicyEvaluation as BasicOPE

# create inputs for OPE
prep = CreateOPEInput(env)
input_dict = prep.obtain_whole_inputs(
    logged_dataset=test_logged_dataset,
    evaluation_policies=evaluation_policies,
    require_value_prediction=True,
    require_weight_prediction=True,  # to estimate marginal importance weights
    n_trajectories_on_policy_evaluation=100,
    random_state=12345,
)

# conduct OPE and visualize the result
ope = BasicOPE(
    logged_dataset=test_logged_dataset,
    ope_estimators=ope_estimators,
)
policy_value_df, policy_value_interval_df = ope.summarize_off_policy_estimates(
    input_dict=input_dict,
    random_state=12345,
)
ope.visualize_off_policy_estimates(
    input_dict, 
    hue="policy", 
    sharey=False,
    random_state=12345,
)
\end{lstlisting}


\begin{figure}[h]
  \centering
  \includegraphics[clip, width=1.0\linewidth]{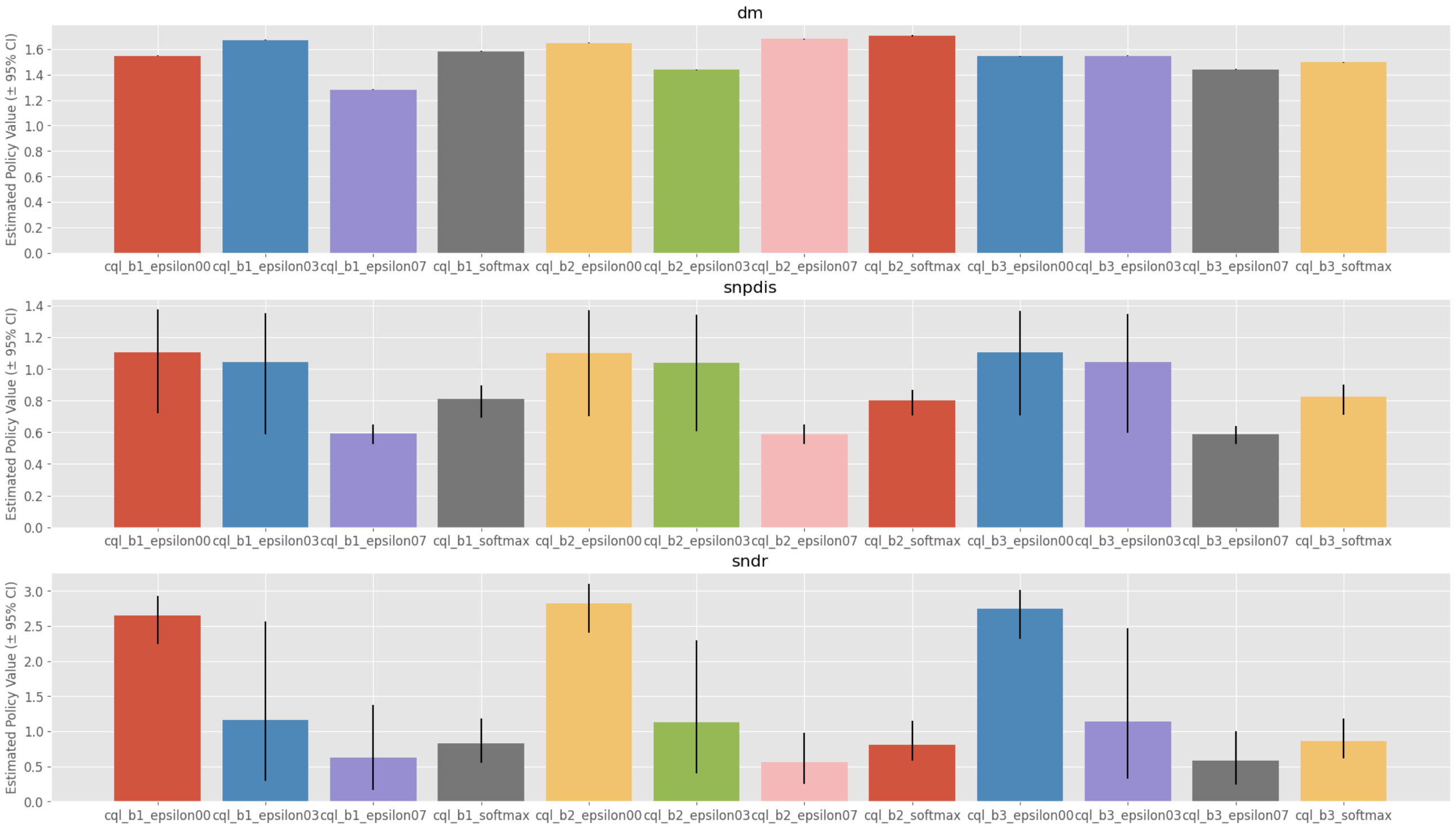}
  \caption{Example of estimating policy value with basic OPE estimators.}
\end{figure}

\newpage

We can also perform CD-OPE in a manner similar to basic OPE as follows.
\begin{lstlisting}[title={Code Snippet 8: \textbf{Cumulative Distribution Off-Policy Evaluation}},captionpos=b]
# import modules and estimators for cumulative distribution OPE from SCOPE-RL
from scope_rl.ope import CumulativeDistributionOPE
from scope_rl.ope.discrete import CumulativeDistributionDM as CD_DM
from scope_rl.ope.discrete import CumulativeDistributionTIS as CD_IS
from scope_rl.ope.discrete import CumulativeDistributionTDR as CD_DR
from scope_rl.ope.discrete import CumulativeDistributionSNTIS as CD_SNIS
from scope_rl.ope.discrete import CumulativeDistributionSNTDR as CD_SNDR

# initialize CumulativeDistributionOPE class
cd_ope = CumulativeDistributionOPE(
    logged_dataset=test_logged_dataset,
    ope_estimators=[CD_DM(), CD_IS(), CD_DR(), CD_SNIS(), CD_SNDR()],
)

# estimate and visualize the cumulative distribution function
cdf_dict = cd_ope.estimate_cumulative_distribution_function(input_dict)
cd_ope.visualize_cumulative_distribution_function(input_dict, n_cols=4)
\end{lstlisting}

\begin{figure}[h]
  \centering
  \includegraphics[clip, width=\linewidth]{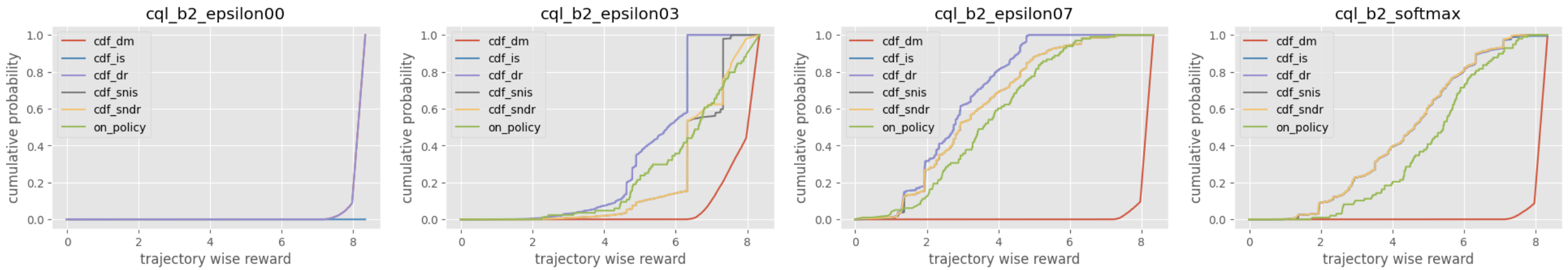}
  \caption{Example of estimating Cumulative Distribution Function (CDF) with CD-OPE estimators.} \label{fig:cdf}
\end{figure}

Similarly, SCOPE-RL is also able to visualize the Conditional Value ar Risk (CVaR) and the interquartile range of the trajectory-wise reward under the evaluation policy as follows.

\begin{figure}[h]
  \centering
  \includegraphics[clip, width=1.0\linewidth]{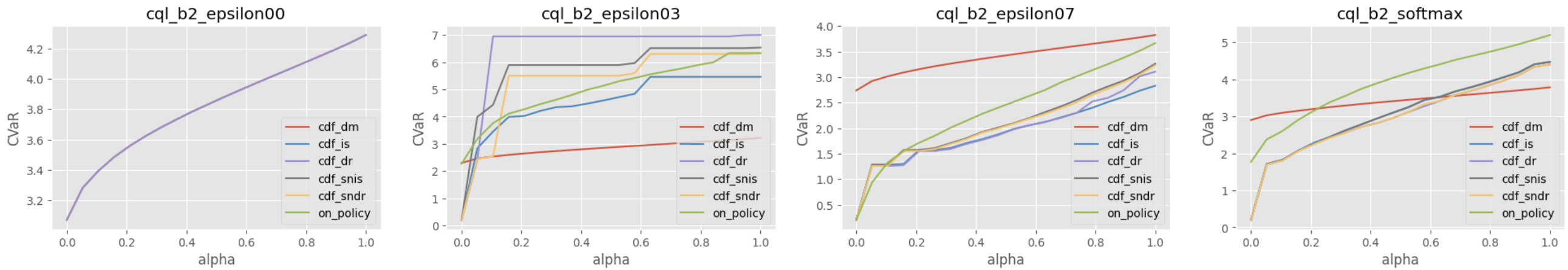}
  \caption{Example of estimating the Conditional Value at Risk (CVaR) with CD-OPE estimators.} \label{fig:cvar}
\end{figure}

\begin{figure}[h]
  \centering
  \includegraphics[clip, width=0.80\linewidth]{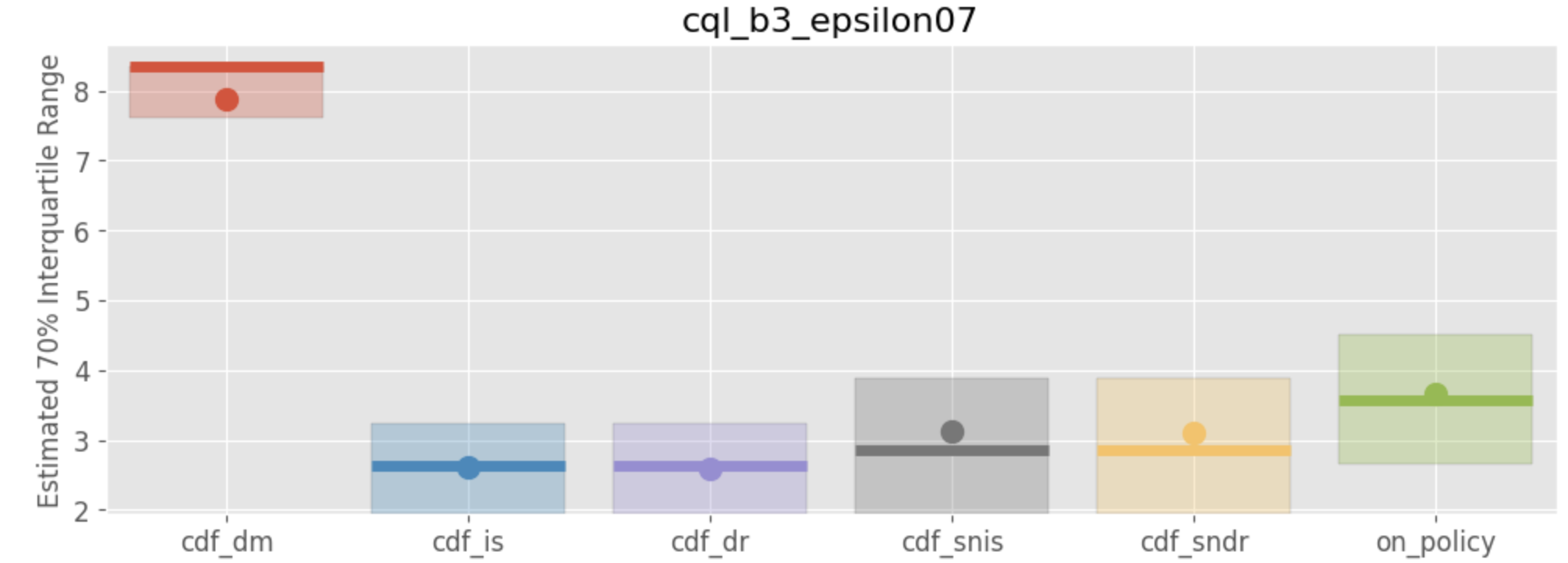}
  \caption{Example of estimating the interquartile range with CD-OPE estimators.} \label{fig:interquartile_range}
\end{figure}

\newpage
\subsubsection{Off-Policy Selection and Assessments of OPE}
Finally, we conduct OPS based on OPE results. We can also evaluate the performance of OPE estimators with SharpRatio@k and other statistics of top-$k$ policies selected by each OPE estimator.
\begin{lstlisting}[title={Code Snippet 9: \textbf{Off-Policy Selection and Assessments of OPE}},captionpos=b]
# import the OPS module from SCOPE-RL
from scope_rl.ope import OffPolicySelection

# initialize the OPS class with OPE instances
ops = OffPolicySelection(
    ope=ope,
    cumulative_distribution_ope=cd_ope,
)

# rank candidate policy by estimated lower quartile and evaluate the OPE results
ranking_df, metric_df = ops.select_by_lower_quartile(
    input_dict,
    alpha=0.3,
    return_metrics=True,
    return_by_dataframe=True,
)

# visualize the top k deployment result
ops.visualize_topk_policy_value_selected_by_standard_ope(
    input_dict=input_dict,
    compared_estimators=["dm", "snpdis", "sndr", "drl"],
    metrics=["best", "worst", "std", "sharpe_ratio"],
    relative_safety_criteria=1.0,
)

# compare the true and estimated policy performances
ops.visualize_policy_value_for_validation(
    input_dict=input_dict,
    compared_estimators=["dm", "snpdis", "sndr", "drl"],
)

\end{lstlisting}

\begin{figure}[h]
  \centering
  \includegraphics[clip, width=1.0\linewidth]{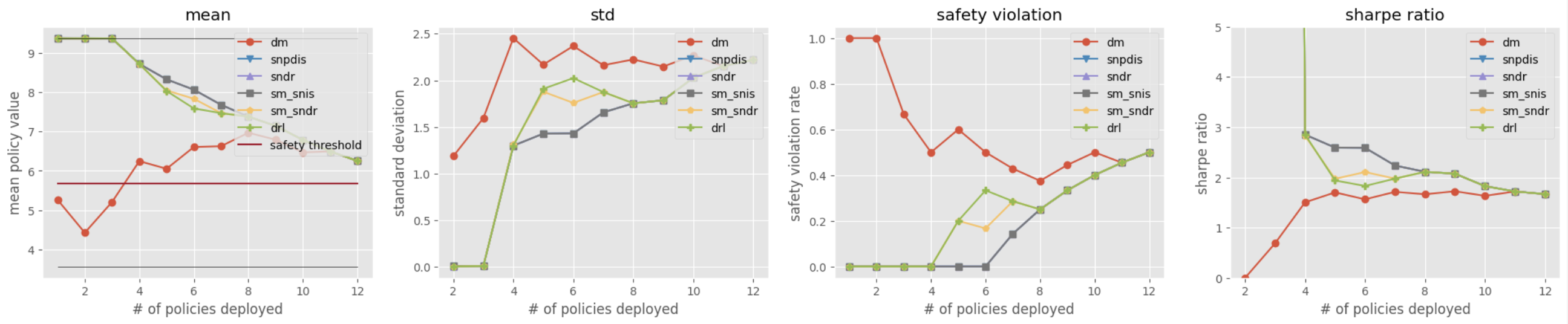}
  \caption{Example of assessing OPE with SharpRatio@k and other statistics of top-$k$ policy portfolio.}
\end{figure}

\begin{figure}[h]
  \centering
  \includegraphics[clip, width=1.0\linewidth]{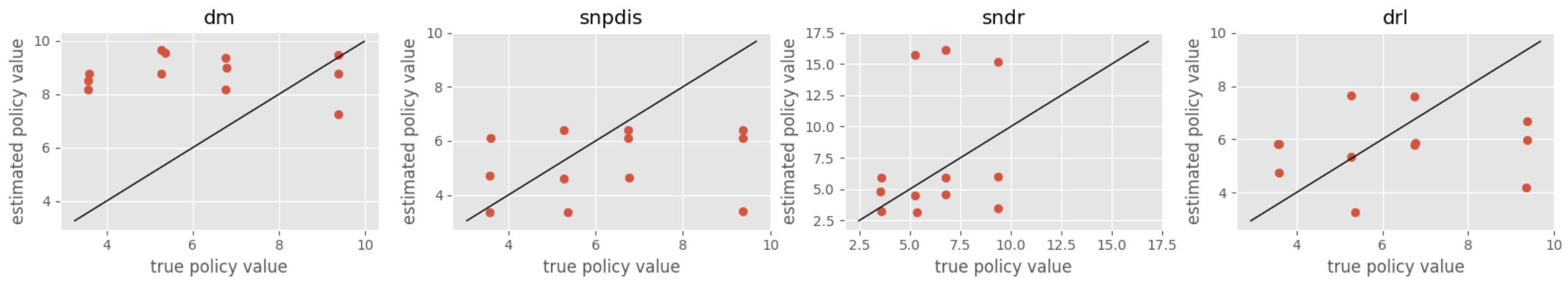}
  \caption{Example of comparing the true and estimated policy value for validation.}
\end{figure}

\newpage

\subsection{Handling multiple logged datasets} \label{app:example_multiple}
SCOPE-RL enables us to conduct OPE and the whole offline RL procedure on multiple logged dataset without additional effort. Below, we show how to handle multiple logged datasets generated by multiple different behavior policies.

\subsubsection{Synthetic Data Generation}
We generate logged datasets with several behavior policies that have different levels of exploration.

\begin{lstlisting}[title={Code Snippet 10: Generating logged datasets with \textit{multiple} behavior policies},captionpos=b]
from scope_rl.dataset import SyntheticDataset
from scope_rl.policy import EpsilonGreedyHead

# define behavior policies
behavior_policies = []
for i, epsilon in enumerate([0.1, 0.3, 0.5]):
    behavior_policy = EpsilonGreedyHead(
        ddqn, 
        n_actions=env.action_space.n,
        epsilon=epsilon,
        name=f"ddqn_eps_{epsilon}",
        random_state=12345,
    )
    behavior_policies.append(behavior_policy)

# initialize dataset class
dataset = SyntheticDataset(
    env=env,
    max_episode_steps=env.step_per_episode,
)

# generate logged datasets by multiple behavior policies
train_logged_dataset = dataset.obtain_episodes(
    behavior_policies=behavior_policies,  #
    n_datasets=10, # number of logged datasets for each behavior policy
    n_trajectories=10000, 
    random_state=12345,
)
test_logged_dataset = dataset.obtain_episodes(
    behavior_policies=behavior_policies,  #
    n_datasets=10, # number of logged datasets for each behavior policy
    n_trajectories=10000, 
    random_state=12345 + 1,
)
\end{lstlisting}

\newpage

\subsubsection{Offline Reinforcement Learning}
To ease the offline policy learning process with multiple logged datasets, SCOPE-RL provides an easy-to-use wrapper for Offline learning. Below, we show the example of obtaining candidate policies with 3 base policies and 2 parameters for exploration as follows.

\begin{lstlisting}[title={Code Snippet 11: Base algorithms and exploration hyperparameters of candidate policies},captionpos=b]
# base algorithms
cql_b1 = CQLConfig(
    encoder_factory=VectorEncoderFactory(hidden_units=[30, 30]),
    q_func_factory=MeanQFunctionFactory(),
).create()
cql_b2 = CQLConfig(
    encoder_factory=VectorEncoderFactory(hidden_units=[100]),
    q_func_factory=MeanQFunctionFactory(),
).create()
cql_b3 = CQLConfig(
    encoder_factory=VectorEncoderFactory(hidden_units=[50, 10]),
    q_func_factory=MeanQFunctionFactory(),
).create()
algorithms = [cql_b1, cql_b2, cql_b3]
algorithms_name = ["cql_b1", "cql_b2", "cql_b3"]

# exploration hyperparameters
policy_wrappers = {
    "eps_03": (
        EpsilonGreedyHead, {
            "epsilon": 0.3,
            "n_actions": env.action_space.n,
        }
    ),
    "eps_05": (
        EpsilonGreedyHead, {
            "epsilon": 0.7,
            "n_actions": env.action_space.n,
        }
    ),
}
\end{lstlisting}

The OPL class trains candidate policies with given algorithms on multiple logged datasets.
\begin{lstlisting}[title={Code Snippet 12: Offline Reinforcement Learning (with \textit{multiple} logged dataset)},captionpos=b]
# import the OPL module from SCOPE-RL
from scope_rl.policy import TrainCandidatePolicies

# initialize the OPL class
orl = TrainCandidatePolicies()

# obtain base policies
base_policies = orl.learn_base_policy(
    logged_dataset=train_logged_dataset,
    algorithms=algorithms,
    random_state=12345,
)
# define evaluation policies
eval_policies = orl.apply_head(
    base_policies=base_policies,
    base_policies_name=algorithms_name,
    policy_wrappers=policy_wrappers,
    random_state=12345,
)
\end{lstlisting}

\newpage

\subsubsection{Off-Policy Evaluation}
SCOPE-RL is also capable of handling OPE with multiple logged datasets in a manner similar to that of a single logged dataset as follows.

\begin{lstlisting}[title={Code Snippet 13: \textbf{Basic Off-Policy Evaluation} (with \textit{multiple} logged datasets)},captionpos=b]
from scope_rl.ope import OffPolicyEvaluation as BasicOPE
from scope_rl.ope import CreateOPEInput
# OPE estimators
from scope_rl.ope.discrete import DirectMethod as DM
from scope_rl.ope.discrete import TrajectoryWiseImportanceSampling as TIS
from scope_rl.ope.discrete import PerDecisionImportanceSampling as PDIS
from scope_rl.ope.discrete import DoublyRobust as DR
from scope_rl.ope.discrete import SelfNormalizedTIS as SNTIS
from scope_rl.ope.discrete import SelfNormalizedPDIS as SNPDIS
from scope_rl.ope.discrete import SelfNormalizedDR as SNDR

# prepare OPE inputs
prep = CreateOPEInput(
    env=env,
)
input_dict = prep.obtain_whole_inputs(
    logged_dataset=test_logged_dataset,  #
    evaluation_policies=eval_policies,
    require_value_prediction=True,
    n_trajectories_on_policy_evaluation=100,
)

# initialize the OPE class with multiple logged datasets
ope = BasicOPE(
    logged_dataset=test_logged_dataset,  #
    ope_estimators=[DM(), TIS(), PDIS(), DR(), SNTIS(), SNPDIS(), SNDR()],
)
# note: the results for each dataset are accessible via ``policy_value_df_dict[behavior_policy.name][dataset_id]``
policy_value_df_dict, policy_value_interval_df_dict = ope.summarize_off_policy_estimates(
    input_dict=input_dict,
    random_state=12345,
)
# conduct OPE with multiple logged datasets and visualize the result
ope.visualize_policy_value_with_multiple_estimates(
    input_dict, 
    hue="policy",
)
\end{lstlisting}

\begin{figure}[h]
  \centering
  \includegraphics[clip, width=0.9\linewidth]{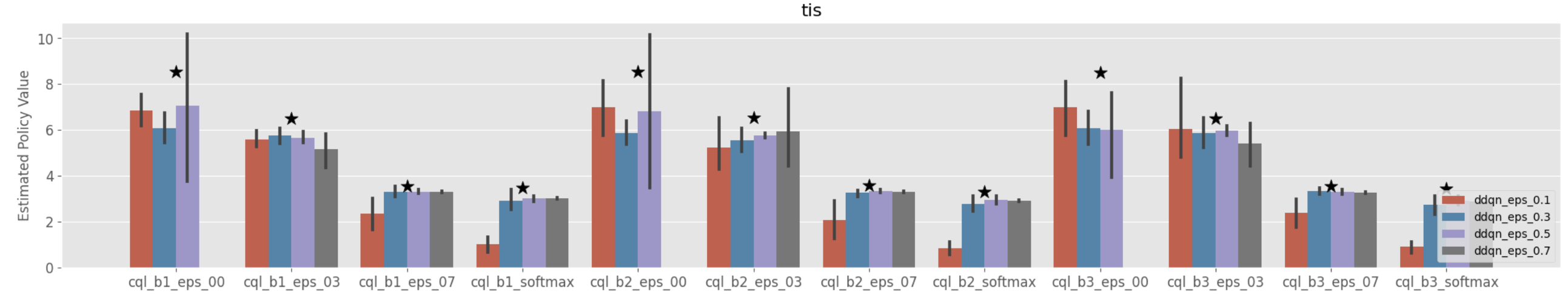}
  \caption{Example of comparing estimated policy value across multiple logged datasets.}
\end{figure}
\newpage

Similarly, SCOPE-RL can handle CD-OPE with multiple logged datasets quite easily.
\begin{lstlisting}[title={Code Snippet 14: \textbf{Cumulative Distribution Off-Policy Evaluation} (with \textit{multiple} logged datasets)},captionpos=b]
from scope_rl.ope import CumulativeDistributionOPE
from scope_rl.ope.discrete import CumulativeDistributionDM as CD_DM
from scope_rl.ope.discrete import CumulativeDistributionTIS as CD_IS
from scope_rl.ope.discrete import CumulativeDistributionTDR as CD_DR
from scope_rl.ope.discrete import CumulativeDistributionSNTIS as CD_SNIS
from scope_rl.ope.discrete import CumulativeDistributionSNTDR as CD_SNDR

# initialize the CD-OPE class with multiple logged datasets
cd_ope = CumulativeDistributionOPE(
    logged_dataset=test_logged_dataset,  #
    ope_estimators=[CD_DM(), CD_IS(), CD_DR(), CD_SNIS(), CD_SNDR()],
)
# note: the results for each dataset are accessible via ``policy_value_dict[behavior_policy.name][dataset_id]``
policy_value_dict = cd_ope.estimate_mean(input_dict)
# estimate and visualize CDF with multiple logged datasets
cd_ope.visualize_cumulative_distribution_function_with_multiple_estimates(
    input_dict,
    behavior_policy_name=behavior_policies[0].name,  # specify behavior policy name
    plot_type="ci_hue",  #
    scale_min=0.0,  # set reward scale (i.e., x-axis or bins of CDF)
    scale_max=10.0,
    n_partition=20,
    n_cols=4,
)

\end{lstlisting}

\begin{figure}[h]
  \centering
  \includegraphics[clip, width=\linewidth]{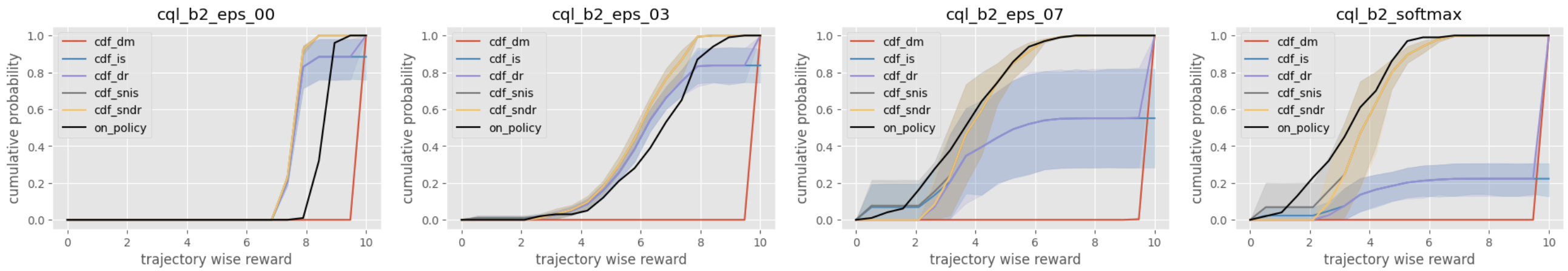}
  \caption{Example of estimating Conditional Value at Risk (CVaR) with CD-OPE estimators on multiple logged datasets.}
\end{figure}

\newpage

\subsubsection{Off-Policy Selection and Assessments of OPE}
Finally, we can also conduct OPS and assess the OPE results with multiple logged datasets as follows.
\begin{lstlisting}[title={Code Snippet 15: \textbf{Off-Policy Selection and Assessmensts of OPE} (with \textit{multiple} logged datasets)},captionpos=b]
from scope_rl.ope import OffPolicySelection

# initialize the OPS class with OPE instances
ops = OffPolicySelection(
    ope=ope,
    cumulative_distribution_ope=cd_ope,
)

# note: the results for each dataset are accessible via ``ranking_df_dict[behavior_policy.name][dataset_id]`` and ``metric_df_dict[behavior_policy.name][dataset_id]``
ranking_df_dict, metric_df_dict = ops.select_by_lower_quartile(
    input_dict,
    alpha=0.3,
    return_metrics=True,
    return_by_dataframe=True,
)

# visualize the top k deployment result
ops.visualize_topk_policy_value_selected_by_standard_ope(
    input_dict=input_dict,
    compared_estimators=["dm", "tis", "pdis", "dr"],
    metrics=["best", "worst", "std", "sharpe_ratio"],
    relative_safety_criteria=1.0,
)

# compare the true and estimated policy performances
ops.visualize_policy_value_for_validation(
    input_dict=input_dict,
    compared_estimators=["dm", "tis", "pdis", "dr"],
)
\end{lstlisting}

\begin{figure}[h]
  \centering
  \includegraphics[clip, width=\linewidth]{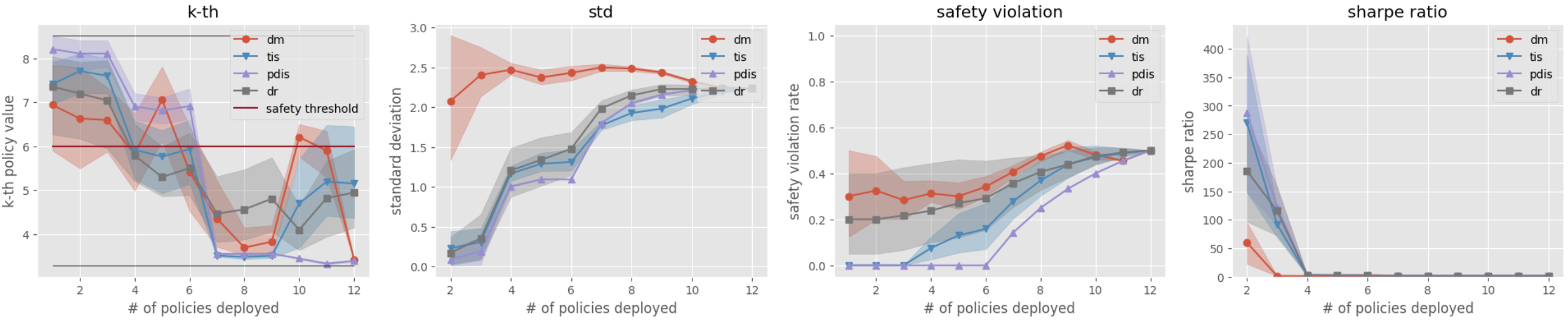}  \caption{Example of assessing OPE with SharpRatio@k and other statistics of top-$k$ policy portfolio with multiple logged datasets.}
\end{figure}

\begin{figure}[h]
  \includegraphics[clip, width=\linewidth]{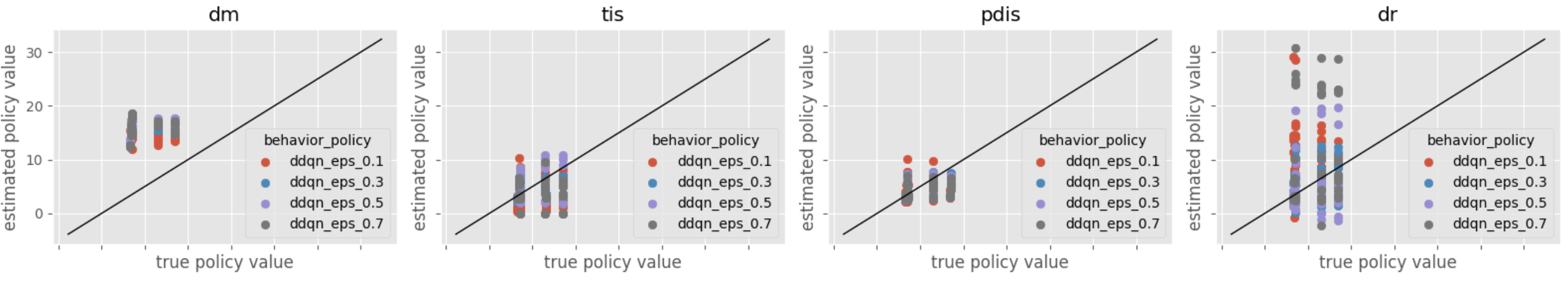}  \caption{Example of comparing the true and estimated policy value for validation with multiple logged datasets.}

\end{figure}

\end{document}